\documentclass{article}

\usepackage{arxiv}

\usepackage{url}
\usepackage{caption}
\usepackage{subcaption}
\usepackage{graphicx}

\usepackage{amsmath}
\renewcommand{\vec}[1]{\boldsymbol{#1}}

\usepackage{acronym}
\newacro{FMU}[FMU]{Functional Mock-up Unit}
\newacroplural{FMU}[FMUs]{Functional Mock-up Units}
\newacro{FMI}[FMI]{Functional Mock-up Interface}
\newacro{ANN}[ANN]{artifical neural network}
\newacroplural{ANN}[ANNs]{artifical neural networks}
\newacro{ODE}[ODE]{ordinary differential equation}
\newacroplural{ODE}[ODEs]{ordinary differential equations}
\newacro{MSL}[MSL]{Modelica Standard Library}
\newacro{PPG}[PPG]{photoplethysmography}
\newacro{CS}[CS]{co-simulation}
\newacro{ME}[ME]{model exchange}
\newacro{PWDB}[PWDB]{Pulse Wave Database}

\newcommand{\myurl}[1]{{\small\url{#1}}}

\newcommand{\libfmi}{{\emph{FMI.jl}}}
\newcommand{\libfmiflux}{{\emph{FMIFlux.jl}}}

\newcommand{\urlfmi}{{\myurl{https://github.com/ThummeTo/FMI.jl}}}
\newcommand{\urlfmiflux}{{\myurl{https://github.com/ThummeTo/FMIFlux.jl}}}

\usepackage{draftwatermark}
\SetWatermarkText{PREPRINT}
\SetWatermarkScale{1}
\definecolor{wm}{gray}{0.95}
\SetWatermarkColor{wm}

\usepackage[dvipsnames]{xcolor}

\title{Hybrid modeling of the human cardiovascular system using NeuralFMUs}

\author{%
	Tobias Thummerer \\
	Chair of Mechatronics\\
	Augsburg University \\
	\texttt{tobias.thummerer@informatik}\\
	\texttt{.uni-augsburg.de}  
	\And 
	Johannes Tintenherr \\
	Chair of Mechatronics\\
	Augsburg University \\
	\texttt{johannes.tintenherr@student}\\
	\texttt{.uni-augsburg.de}  
	\And 
	Lars Mikelsons \\
	Chair of Mechatronics\\
	Augsburg University \\
	\texttt{lars.mikelsons@informatik}\\
	\texttt{.uni-augsburg.de}  
}
\date{}

\renewcommand{\headeright}{Preprint}
\renewcommand{\undertitle}{Preprint}
\renewcommand{\shorttitle}{~}

\begin{document}

\maketitle

\begin{abstract}
Hybrid modeling, the combination of first principle and machine learning models, is an emerging research field that gathers more and more attention. Even if hybrid models produce formidable results for academic examples, there are still different technical challenges that hinder the use of hybrid modeling in real-world applications. By presenting \emph{Neural\acsp{FMU}}, the fusion of a \ac{FMU}, a numerical \acs{ODE} solver and an \acl{ANN}, we are paving the way for the use of a variety of first principle models from different modeling tools as parts of hybrid models. This contribution handles the hybrid modeling of a complex, real-world example: Starting with a simplified 1D-fluid model of the human cardiovascular system (arterial side), the aim is to learn neglected physical effects like arterial elasticity from data. We will show that the hybrid modeling process is more comfortable, needs less system knowledge and is therefore less error-prone compared to modeling solely based on first principle. Further, the resulting hybrid model has improved in computation performance, compared to a pure first principle white-box model, while still fulfilling the requirements regarding accuracy of the considered hemodynamic quantities. The use of the presented techniques is explained in a general manner and the considered use-case can serve as example for other modeling and simulation applications in and beyond the medical domain.
\end{abstract}

\section{Introduction}
The structural integration of physical white-box models into \acp{ANN} to retrieve a hybrid model is a growing research field, see \cite{Willard:2020} or \cite{Rai:2020}. One milestone was the use of state-of-the-art numerical solvers for \acp{ODE} inside of \acp{ANN} instead of residual net structures to reproduce numerical integration and learn dynamic system behavior, like in \cite{Chen:2018}. The resulting combination of an \ac{ANN} and an \ac{ODE} solver - so called \emph{NeuralODEs} - lead to improvements in model accuracy while also enhancing computation and memory cost. 

Based on the idea of Neural\acsp{ODE} we presented \emph{Neural\acsp{FMU}} in \cite{Thu:2021}, the structural integration of a first-principle model in form of a \ac{FMU}, a modern numerical solver like \emph{Tsit5} \cite{Tsitouras:2011} or \emph{CVODE} \cite{CVODE:2021} and a feed-forward \ac{ANN}. Further, we provided the open-source frameworks \libfmi{}\footnote{\urlfmi{}} and \libfmiflux{}\footnote{\urlfmiflux{}} to allow for the setup and training of Neural\acp{FMU} just like a convenient \ac{ANN} in the Julia programming language. In this contribution, the considered technologies shall be tested with a model of the human cardiovascular system (s. Sec. \ref{sec:refmodel}). 

One of the core challenges in hybrid modeling, the fusion of data driven and first principle models, is to provide an efficient training process for the resulting heterogeneous structure. For training of \acp{ANN} in general, the gradient of the loss function according to the net parameters is needed. This belongs also to hybrid models: The gradient must be determined along the \ac{ANN}, the numerical solver and the model of the physical system. The different jacobians must be determined by different methods, because of the availability (or non-availability) of information and interfaces between the Neural\ac{FMU} components. For example, the mathematical operations inside of \acp{ANN} are known, therefore many methods to retrieve partial derivatives, like automatic differentiation, are possible to use. Models exported from modeling tools on the other hand need different techniques, because they do not necessarily provide the symbolic structure of the model or often actively hide this information because it may contain sensible company knowledge. For \acp{FMU}, we suggest the use of the optional built-in function \verb"fmi2GetDirectionalDerivative" or - if not available - sampling of additional simulation points and partial derivative approximation via finite (central) differences. Both approaches are implemented as part of \libfmiflux{}. Finally, if all jacobians over every component are retrieved, the needed differentiation chains must be deployed to the machine learning framework. For an overview and more detailed insight into the necessary technical steps, see \cite{Thu:2021}.
\\
\\
In the following, some short style explanations about involved tools and standards are given.

\subsection{Julia programming language}
The Julia Programming Language (hereinafter: \emph{Julia}) is a dynamic typing language, being developed at the \emph{Massachusetts Institute of Technology} since 2009 and first published in 2012 \cite{Bezanson:2012}. Julia provides the ability to run code platform-independent and with performance benchmarks similar to native C-implementations. Regardless of the fact that Julia is relatively young (in terms of programming languages), the community is growing rapidly and many other research facilities joined the development of the language and language extensions. 

\subsection{Modelica}
The physical modeling was performed in the object-orientated modeling language \emph{Modelica} (\myurl{https://modelica.org/}). Modelica allows acausal modeling, meaning causalization of the system of equations is handled automatically by the compiler at compile time. This allows for the building of large models, while keeping them human understandable by using an intuitive topology and sub-model enclosures. The two most common tools for graphical supported programming with Modelica are Dymola (by Dassault Systèmes \textregistered{}) and OMEdit (open-source), both tools allow for  model export as \ac{FMU} (s. next subsection).

\subsection{\acl{FMI}}
\ac{FMI} allows for the simulation and parameterization of models outside of the original modeling environment in a standardized and platform-independent format. It is possible to generate standard-compliant models, so called \acp{FMU}, in two different modes: \ac{ME} and \ac{CS}. \ac{ME}-\acp{FMU} offer an interface for the system dynamics, meaning the \ac{FMU} computes a system state derivative for a given system state. In a subsequent step outside of the \ac{FMU}, the next system state can be derived by numerically integrating the state derivative. \ac{CS}-\acp{FMU} already include a numerical \ac{ODE} solver, which allows for an even easier simulation. On the other hand, this inclusion prevents manipulation of the system dynamics before the numerical integration and consequently makes this mode unattractive for the aim of this contribution.

This paper further divides into three sections: The presentation of the cardiovascular system model, modeling assumptions and the extension to a hybrid model (s. Sec. \ref{sec:model}), followed by training and validation of the hybrid model (s. Sec. \ref{sec:training}) and finally a short conclusion, including current and future work (s. Sec. \ref{sec:conclusion}).

\section{Modeling}\label{sec:model}
\subsection{Reference model}\label{sec:refmodel}
In \cite{Charlton:2019}, a model to simulate arterial pulse waves and their changes over aging of healthy patients is introduced. Together with the model itself, simulation results for arterial blood pressure, volume flow, cross section (luminal area) and \ac{PPG} for 4,374 model patients of ages between 25 and 75 years are published. The model and simulation data was validated with in vivo data from different sources. The fluid simulation is implemented as laminar, incompressible, newtonian 1D-flow, extended by a viscoelastic term to model diameter changes in arterial cross sections. The simulation setup and parameterization was done in \emph{Matlab} (by MathWorks \textregistered{}), the 1D-fluid simulation was performed with \emph{Nektar1D} (\myurl{http://haemod.uk/nektar}) using the hemodynamic model from \cite{Alastruey:2012}. The arterial system is closed by boundary conditions: The heart is abstracted as predetermined mass flow into the Aorta, the vascular beds (the remaining boundary conditions of the fluid simulation) where modeled as three-element windkessels. Please note, that only the \emph{arterial side} of the cardiovascular system is modeled: Starting at the heart, the blood flow passes multiple arteries and finally reaches the vascular beds.

\begin{figure}[h!]
	\centering
	\includegraphics[width=14cm]{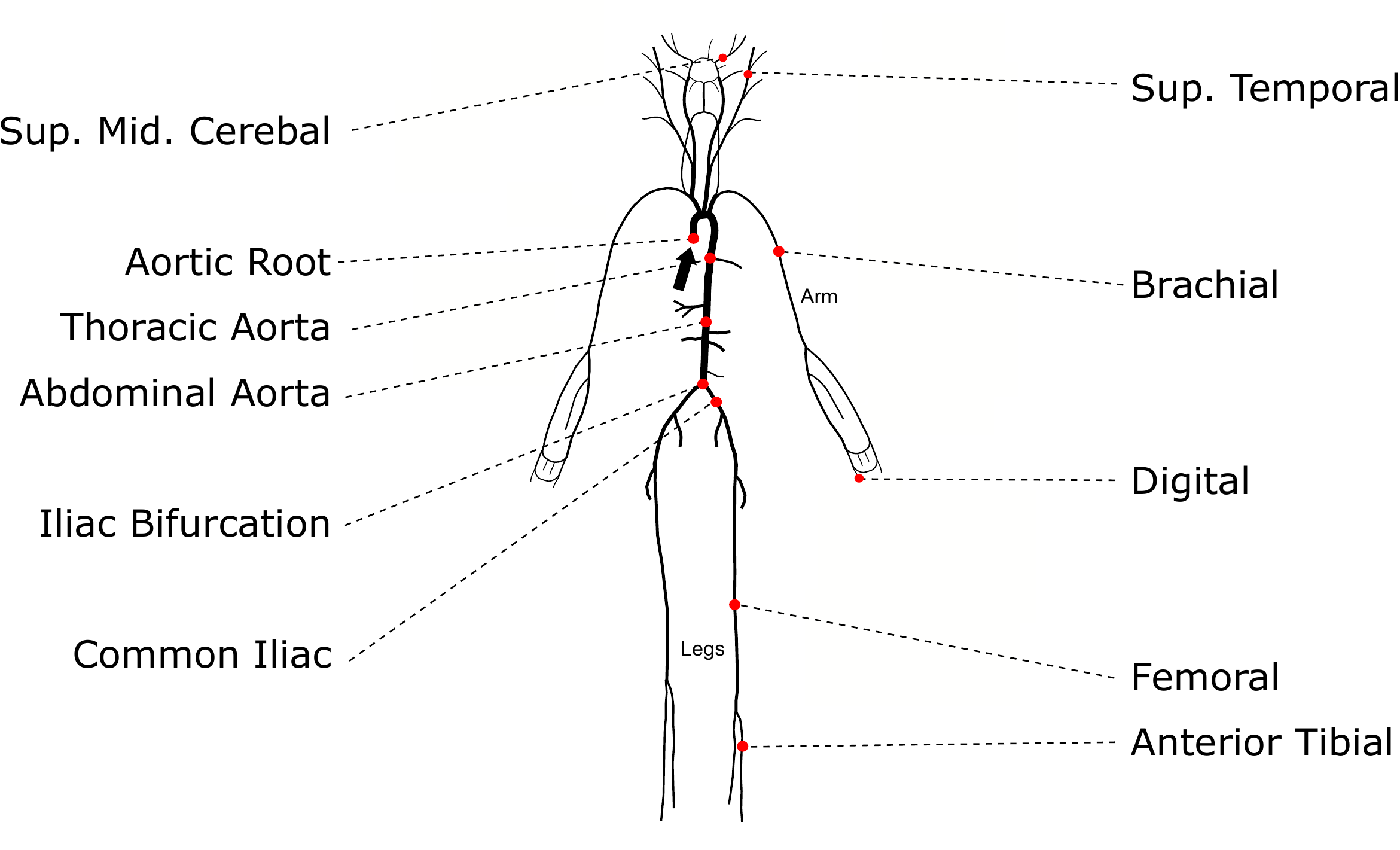}\hspace{2pc}%
	\caption{The simplified structure of the arterial side of the human cardiovascular system (figure adapted from \cite{Charlton:2019}). The hemodynamics at the 10 selected artery segments (red) will be learned from data.}
	\label{fig:arteries}
\end{figure}

The simulation database of the introduced contribution, further referenced as \ac{PWDB} (\myurl{https://zenodo.org/record/3275625}), serves as an ideal starting point for the presented machine learning task. The \ac{PWDB} includes ready-to-use data for 12 arterial segments, the following 10 segments of this set will be used for the later training process:

\begin{itemize}
	\item \emph{Thoracic Aorta}, end of segment 18
	\item \emph{Abdominal Aorta}, start of segment 39
	\item \emph{Iliac Bifurcation}, end of segment 41
	\item left \emph{Superficial Temporal Artery}, end of segment 87
	\item left \emph{Superior Middle Cerebral Artery}, end of segment 72
	\item left \emph{Brachial Artery}, three quarters along segment 21
	\item left \emph{Digital Arteries}, end of segment 112
	\item left \emph{Iliac Artery}, half way of segment 44
	\item left \emph{Femoral Artery}, half way of segment 46
	\item left \emph{Tibial Artery}, end of segment 49
\end{itemize}

\subsection{First principle model (Modelica)}
We start by building up a first principle model in form of an intuitive, object-orientated Modelica model, using default components from the \ac{MSL} as far as possible. However, the presented procedure is not limited to Modelica models, of course.

\begin{figure}[h!]
	\centering
	\includegraphics[width=\textwidth]{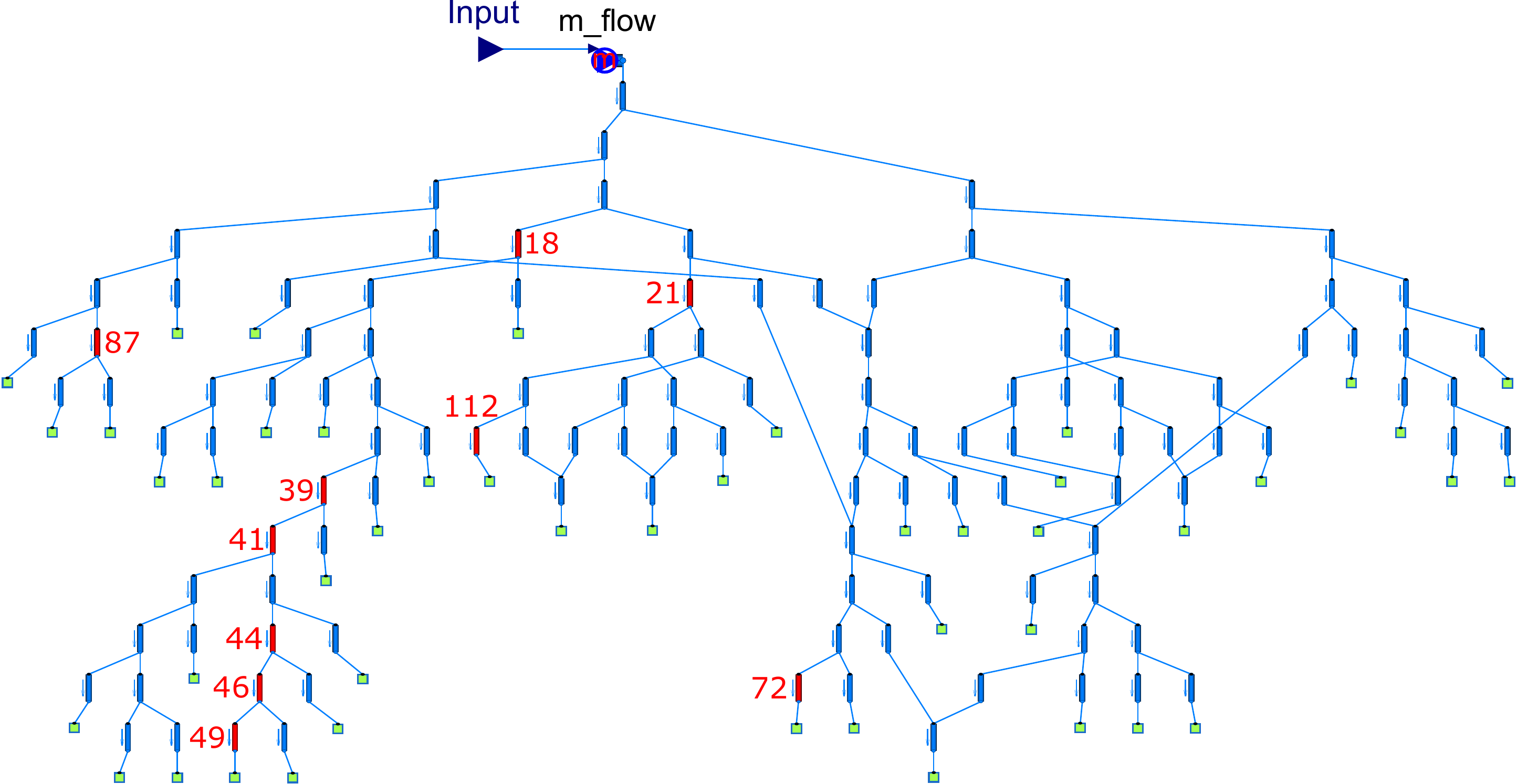}\hspace{2pc}%
	\caption{The physical model of the human cardiovascular system (arterial side), consisting of 116 static pipes (arterial segments, red/blue) and 46 three-element windkessels (vascular beds, green), in \emph{Dymola}. From the set of pipes, 10 segments were picked for the machine learning task (red). The mass flow from heart into the Aorta \emph{m\_flow} is given as model input. The model was build with components of the \ac{MSL}.}
	\label{fig:dymola}
\end{figure}

\newpage
The first principle model of the cardiovascular system is a simplification of the reference model concerning the following aspects:
\begin{itemize}
\item The arteries are modeled as static pipes (\ac{MSL} 3.2.3: Modelica.Fluid.Pipes.StaticPipe), meaning mass, momentum and energy balances are assumed steady-state. As a result, the pipe itself does not store any mass or energy, but physical conservation principles are satisfied.	
\item The diameters of the arterial segments are assumed constant, meaning the arterial cross section does not vary dependent on the system state (pressure).
\item Finally, the arteries are assumed cylindrical, so the inlet and outlet radius are the same size. To parameterize the pipes with parameter data from the reference model, which assumes a pipe with constant diameter change over length, the mean value of inlet and outlet diameter was used.
\end{itemize}

\begin{figure}[h!]
	\centering

	\begin{subfigure}[b]{0.45\textwidth}
		\centering
		\includegraphics[width=\textwidth]{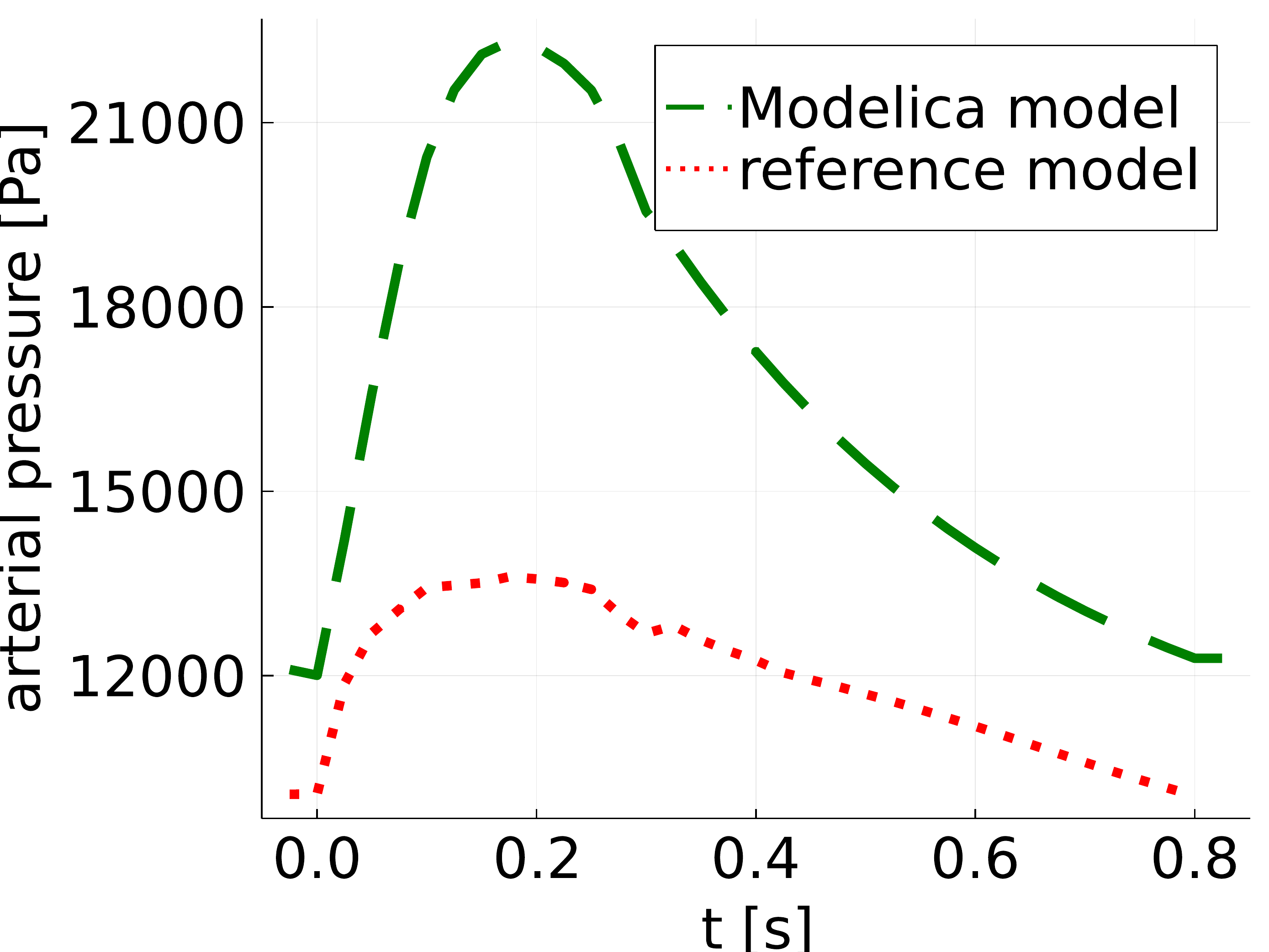}
		\caption{Thoracic Aorta (\#18)}
	\end{subfigure}
	\hfill
	\begin{subfigure}[b]{0.45\textwidth}
		\centering
		\includegraphics[width=\textwidth]{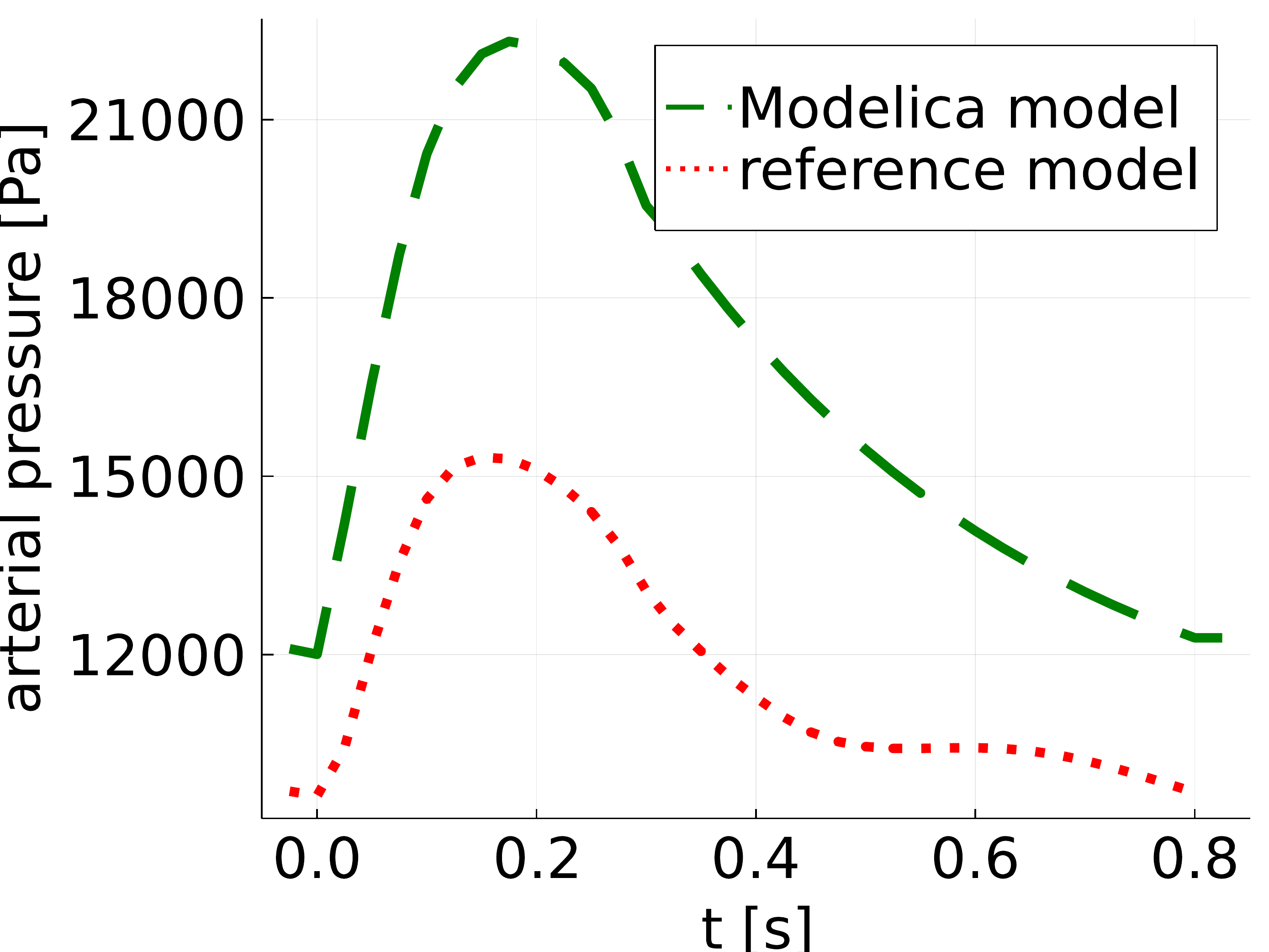}
		\caption{Digital Arteries (\#112)}
	\end{subfigure}
	
	\caption{Deviation between the Modelica model (green/dashed) and the reference model (red/dotted) for the simulation of a single pressure pulse wave of subject \#1, observed at two different segments.}
	\label{fig:deviation}
\end{figure}

As to expect, these modeling simplifications lead to a significantly different simulation result compared to the reference model (s. Fig \ref{fig:deviation}). However, the pulse wave characteristic is roughly visible despite the simplifications. The main reason for the difference between reference and simplified Modelica model is the lack of arterial dynamics: The system pressure inside the simplified model is the same at any location inside the arterial network, but varies over time with the heart pulse wave. On the other hand, the simplified model has a much better computational performance. The motivation for hybrid modeling is to retrieve model accuracy close to the reference model, but providing better computational performance at the same time.

To conclude, the only remaining system dynamics are the pressures over the 46 terminal windkessels, meaning the system state $\vec{x}_{wk}$ (\emph{\textbf{w}ind\textbf{k}essels}) can be uniquely described using a 46 entries state vector. To retrieve a more exact model that describes a dynamic pressure drop over the arterial segments dependent on arterial cross section change by elasticity, more states are needed. To setup the Modelica model for machine learning, placeholder states $\vec{x}_{art}$ (\emph{\textbf{art}eries}) were inserted into the system at the considered artery locations, resulting in the concatenated system state vector $\vec{x} = \vec{x}_{wk} | \vec{x}_{art}$. This state space expansion occurs by simply inserting components into the object-orientated model at the desired locations. Two placeholders types are examined: The insertion of fluid capacities with small capacitance (hereinafter: \emph{C-placeholders}, one state) and the insertion of parallel circuits consisting of a capacitor and an inductance (hereinafter: \emph{LC-placeholders}, two states). This reflects a straightforward model enhancement process: Adding dynamic placeholders to a model at the locations needed, while the correct physical term is learned from data. Note, that the use of a variety of placeholders is possible, like RLC-circuits for example. After this final step, both models (one with C- and one with LC-placeholders) were exported as \acp{FMU}.

\subsection{Hybrid model}
Hybrid modeling is a technique picked often, if higher modeling accuracy is needed, but conventional first principle modeling does not increase precision, is not economically practicable anymore or is simply not possible because of lack of system knowledge. On the other hand, almost any development process in medicine, mechanics or electronics generates data, that can be used to achieve better (more precise or faster computing) hybrid models.

During hybrid modeling, we will neglect knowing the reference system and view the problem from a different perspective: Starting with the simple Modelica model, we want to improve model accuracy without knowing the physical principles needed for this step. Therefore, the improvement process is done solely on basis of the reference model simulation data, not on physical system knowledge. This mirrors a typical use case: Model accuracy shall be improved, further first principle modeling is not an option, but measurement data of a more accurate (or the real) system is available and can be used.

\begin{figure}[h!]
	\centering 
	\includegraphics[height=\textheight]{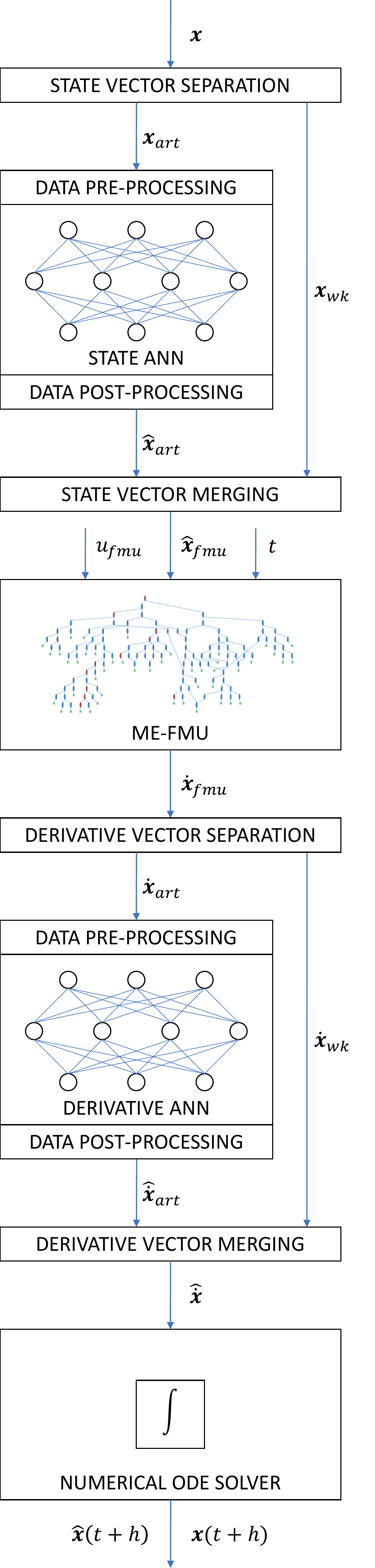}\hspace{2pc}%
	\begin{minipage}[b]{12pc}
		\caption{
			The hybrid model structure. From top to bottom: Because not all states of the NeuralFMU shall be part of the training process, the state vector $\vec{x}$ is separated into the two sub-vectors $\vec{x}_{wk}$ (the windkessels' states) and $\vec{x}_{art}$ (the arterial pressure states). Before being fed into the upper \ac{ANN}, the arterial state values $\vec{x}_{art}$ are pre-processed to ensure being inside an \ac{ANN}-compliant range and are transformed back after the \ac{ANN} pass. The state \ac{ANN} at the top of the topology is able to learn and compensate state dependent modeling failures, like static offsets in measurement data in this special case.
			\\
			The new arterial state estimate $\hat{\vec{x}}_{art}$ is merged together with the original windkessel states $\vec{x}_{wk}$ into the modified system state vector $\hat{\vec{x}}_{fmu}$, that is further passed, together with the system input $u_{fmu}$ and the current simulation time $t$, to the \ac{ME}-\ac{FMU}.
			\\
			Inside the \ac{FMU}, the current state derivatives $\dot{\vec{x}}_{fmu}$ are computed based on the given system state $\hat{\vec{x}}_{fmu}$.
			\\
			As for the system state, the system state derivative vector is separated into the arterial $\dot{\vec{x}}_{art}$ and windkessel state derivative vector $\dot{\vec{x}}_{wk}$. The arterial pressure dynamics are pre-processed, fed to the \ac{ANN}, post-processed and the resulting $\hat{\dot{\vec{x}}}_{art}$ is merged back together with $\dot{\vec{x}}_{wk}$ into the system state derivative estimate $\hat{\dot{\vec{x}}}$. The derivative \ac{ANN} is therefore able to manipulate the system dynamics to retrieve a different system behavior. In the considered use case, the pressure drop over the arterial segments can be learned based on data. Because the \ac{ANN} is learning on basis of state derivatives, generalized learning of the correct physical laws is promoted.
			\\
			Finally, the system dynamics estimate $\hat{\dot{\vec{x}}}$ is numerically integrated with step size $h$ into the next system state estimate $\hat{\vec{x}}(t+h)$ or the next system state $\vec{x}(t+h)$ respectively.
		}
		\label{fig:structure}
	\end{minipage}
\end{figure}

To start with hybrid modeling, the \ac{FMU} models exported from Dymola are imported into Julia using the library \libfmi{}. With the library extension \libfmiflux{}, we can setup and train Neural\acp{FMU} on basis of the Modelica model \acp{FMU}. For the considered use-case, two Neural\acp{FMU} as in Fig. \ref{fig:structure} (one with C- and one with LC-placeholders) with layer dimensions as in Tab. \ref{tab:layout} were examined.

\begin{center}
	\begin{table}[h!]
		\caption{\label{tab:layout}Topologies of the considered Neural\ac{FMU}s.}
		\centering
		\begin{tabular}{llccccl}
			\hline
			Layer & Type & \multicolumn{2}{c}{C-placeholders} & \multicolumn{2}{c}{LC-placeholders} & Activation \\
			~ & ~ & Inputs & Outputs & Inputs & Outputs & ~ \\
			\hline
			\#1 & state vector separation & 56 & 10 $|$ 46 & 66 & 20 $|$ 46 & none \\
			\#2 & data pre-processing & 10 & 10 & 20 & 20 & none \\
			\#3 & bias & 10 & 10 & 20 & 20 & none \\
			\#4 & data post-processing & 10 & 10 & 20 & 20 & none \\
			\#5 & state vector merging & 10 $|$ 46 & 56 & 20 $|$ 46 & 66 & none\\
			\#6 & \ac{FMU} & 56 & 56 & 66 & 66 & none\\
			\#7 & derivative vector separation & 56 & 10 $|$ 46 & 66 & 20 $|$ 46 & none \\
			\#8 & data pre-processing & 10 & 10 & 20 & 20 & none \\
			\#9 & dense & 10 & 30 & 20 & 30 & tanh \\
			\#10 & dense & 30 & 10 & 30 & 20 & none \\
			\#11 & data post-processing & 10 & 10 & 20 & 20 & none \\
			\#12 & derivative vector merging & 10 $|$ 46 & 56 & 20 $|$ 46 & 66 & none\\
			\hline
		\end{tabular}
	\end{table}
\end{center}

\section{Training \& Validation}\label{sec:training}

During training, three pulse wave cycles for patient \#1 are simulated ($2.466\,s$, varies between subjects because of different heart rates). The first pulse wave is ignored in the training process and allows the system to retrieve a stationary (periodically repeating) state. The training takes place on two cycles to promote learning a time-invariant system behavior, meaning a periodically repeating input (heart) should generate a periodically repeating output (arterial pressure curves) with same period length. Even if the \ac{PWDB} includes data observed at $500\,Hz$, the training process is sampled down to $40\,Hz$ because of training performance optimization. To conclude, training takes place on only two pulse waves of one patient, resulting in 66 data points à 10 arterial pressure values. Note, that this is a very small training base in the field of machine learning. 

\begin{figure}[h!]
	\centering
	
	\begin{subfigure}[b]{0.32\textwidth}
		\centering
		\includegraphics[width=\textwidth]{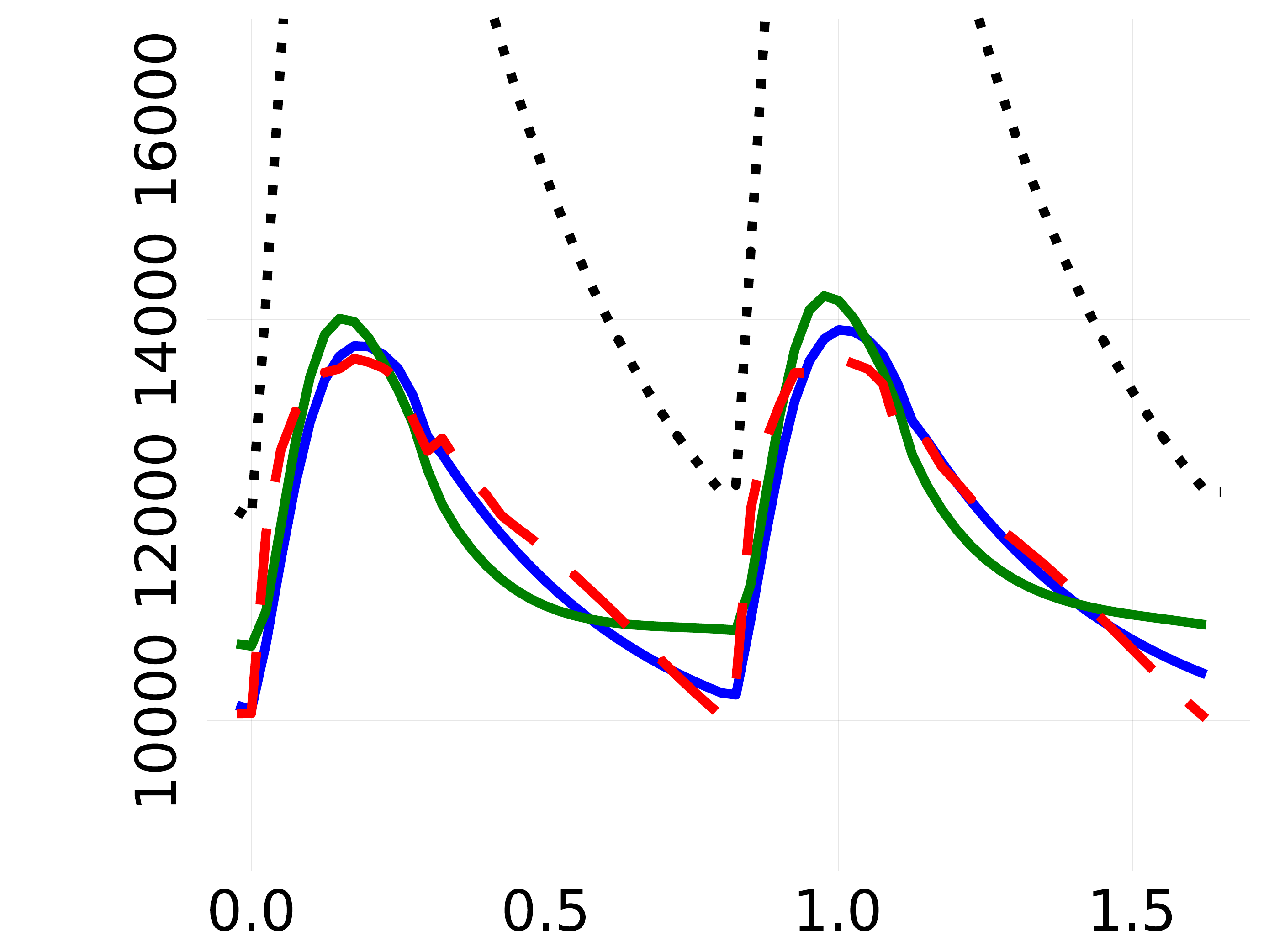}
		\caption{Thoracic Aorta (\#18)}
	\end{subfigure}
	\hfill
	\begin{subfigure}[b]{0.32\textwidth}
		\centering
		\includegraphics[width=\textwidth]{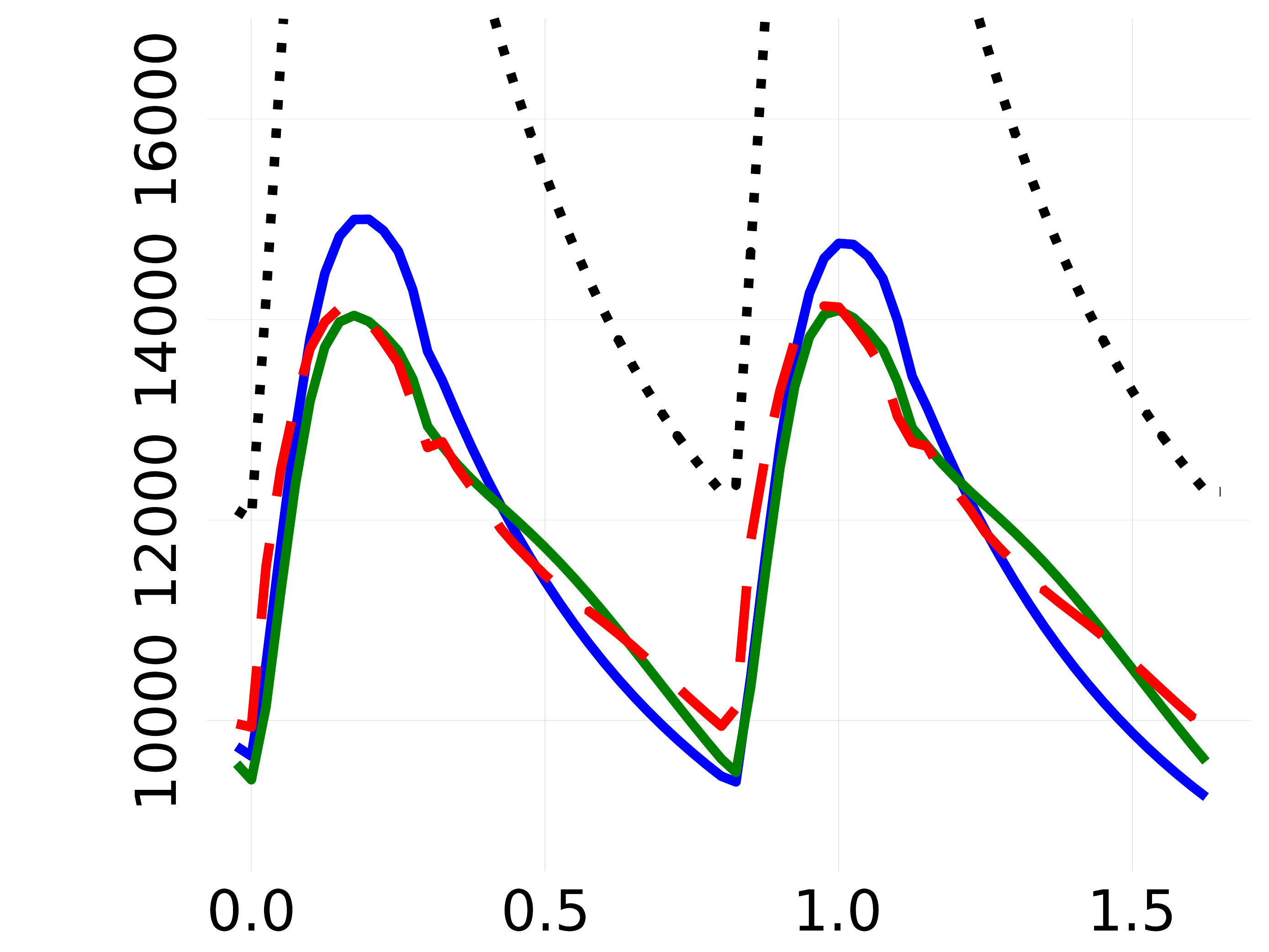}
		\caption{Abdominal Aorta (\#39)}
	\end{subfigure}
	\hfill
	\begin{subfigure}[b]{0.32\textwidth}
		\centering
		\includegraphics[width=\textwidth]{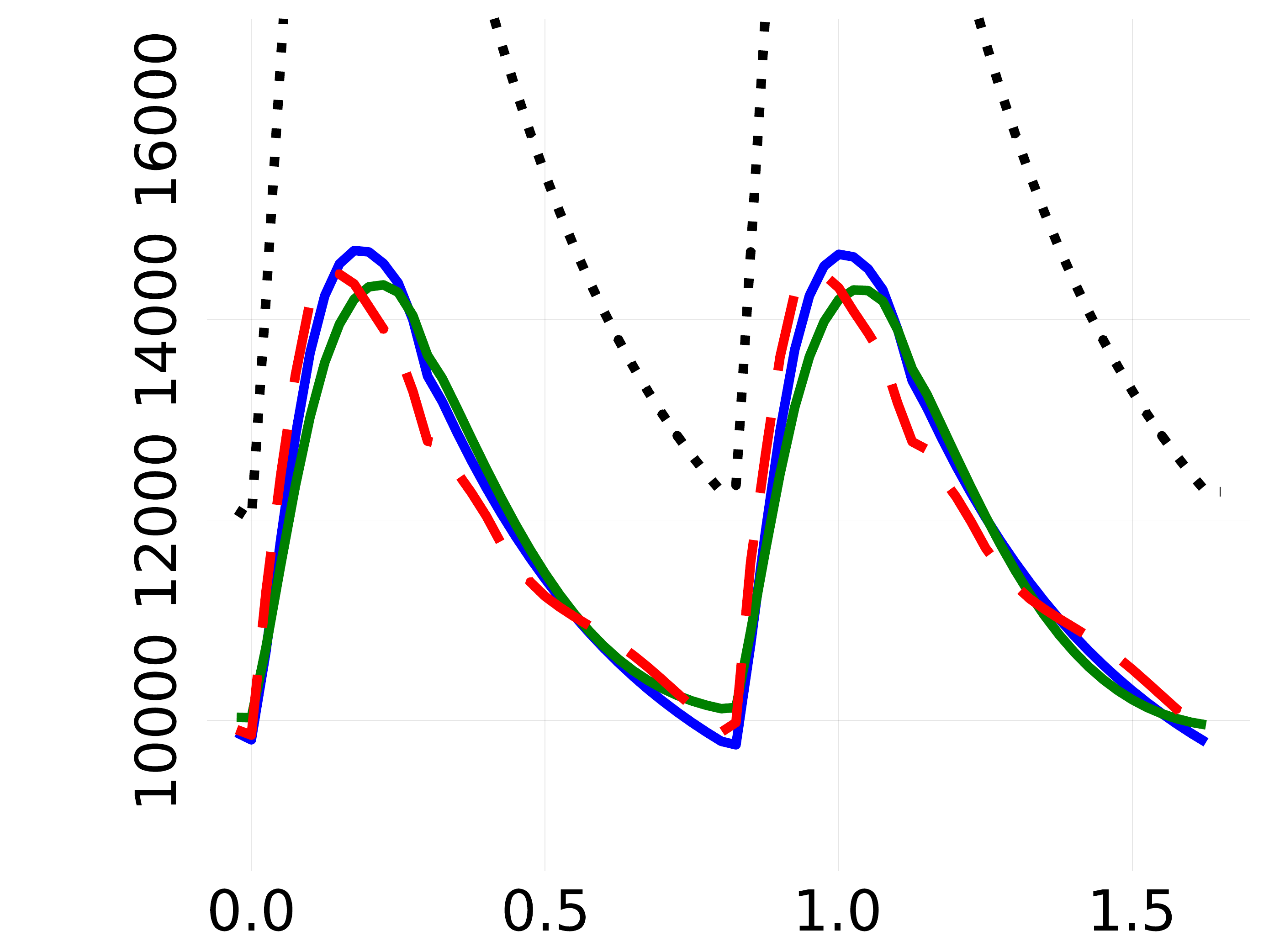}
		\caption{Iliac Bifurcation (\#41)}
		\label{fig:five over x}
	\end{subfigure}

	\vspace{0.5cm}
	\begin{subfigure}[b]{0.32\textwidth}
		\centering
		\includegraphics[width=\textwidth]{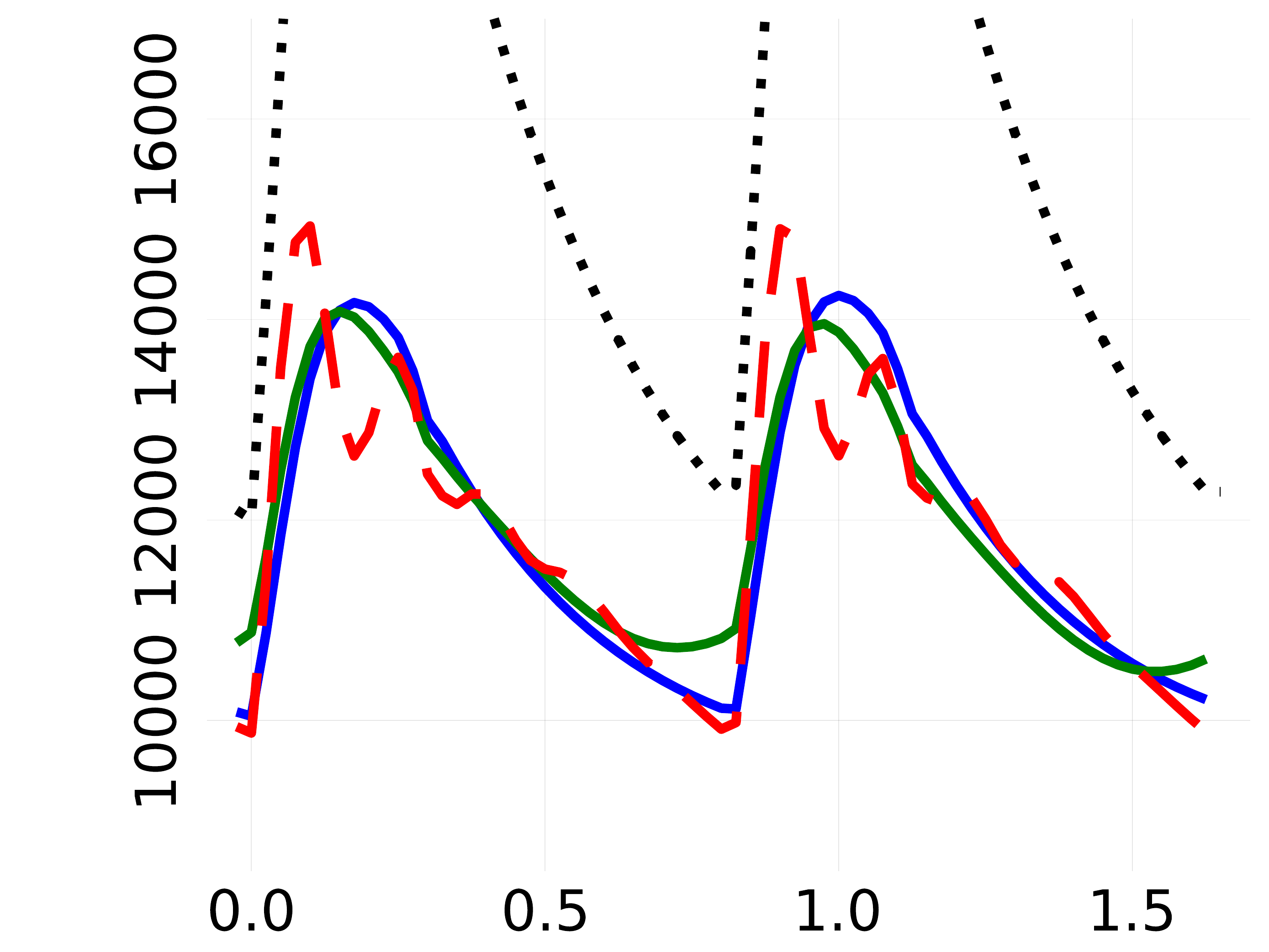}
		\caption{Sup. Temp. Artery (\#87)}
	\end{subfigure}
	\hfill
	\begin{subfigure}[b]{0.32\textwidth}
		\centering
		\includegraphics[width=\textwidth]{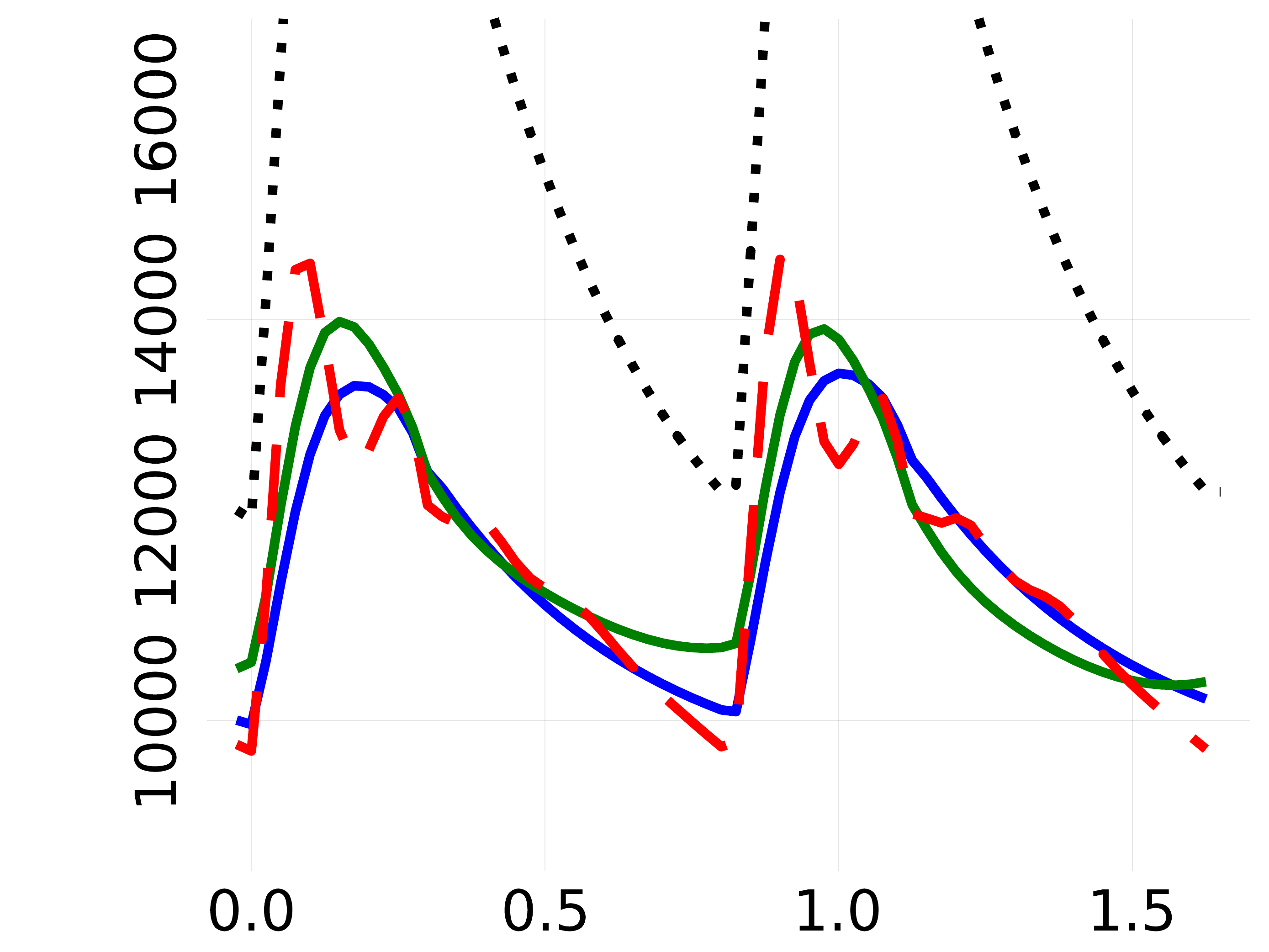}
		\caption{Mid. Cerebral Artery (\#72)}
	\end{subfigure}
	\hfill
	\begin{subfigure}[b]{0.32\textwidth}
		\centering
		\includegraphics[width=\textwidth]{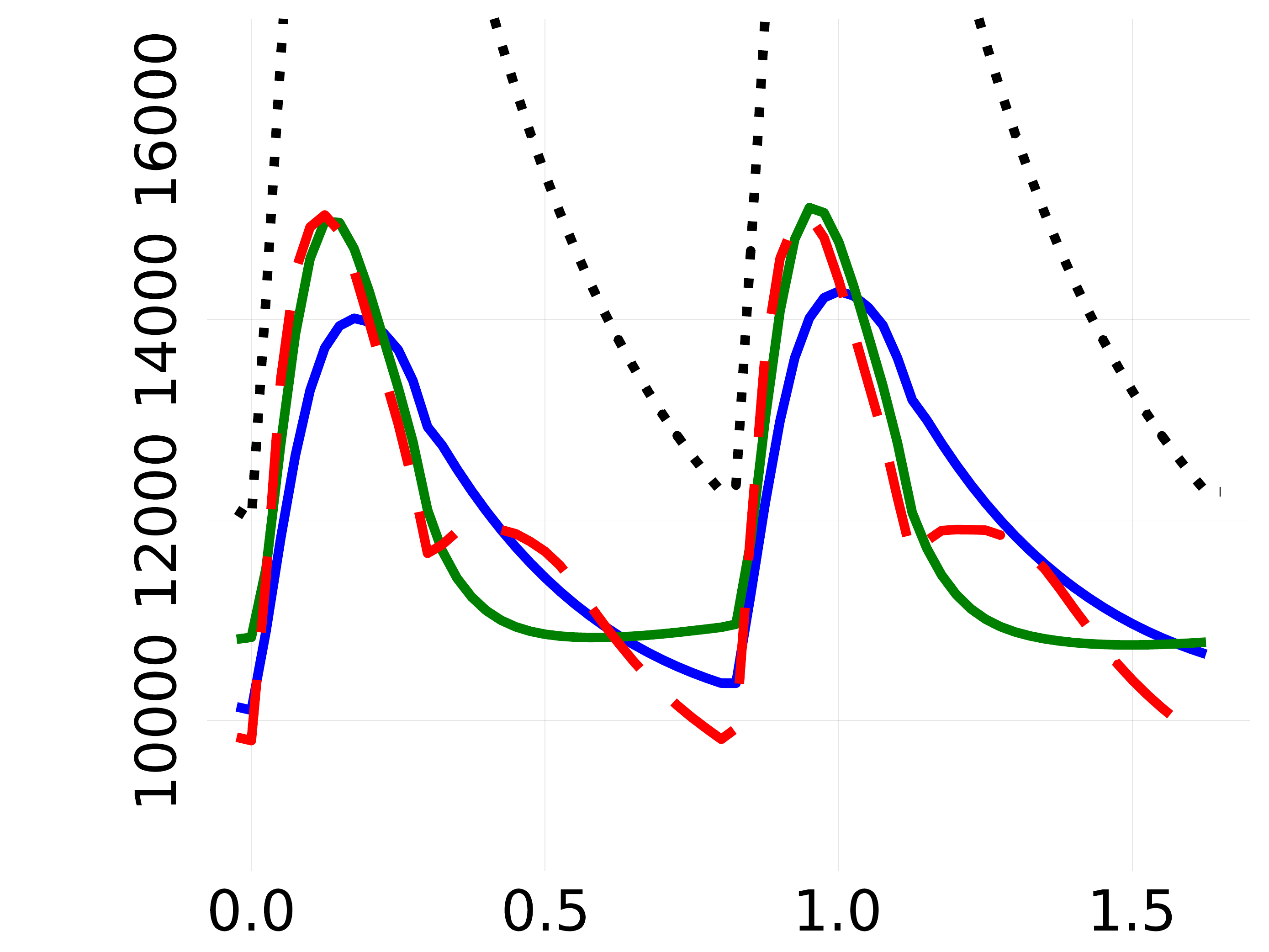}
		\caption{Brachial Artery (\#21)}
	\end{subfigure}

	\vspace{0.5cm}
	\begin{subfigure}[b]{0.32\textwidth}
		\centering
		\includegraphics[width=\textwidth]{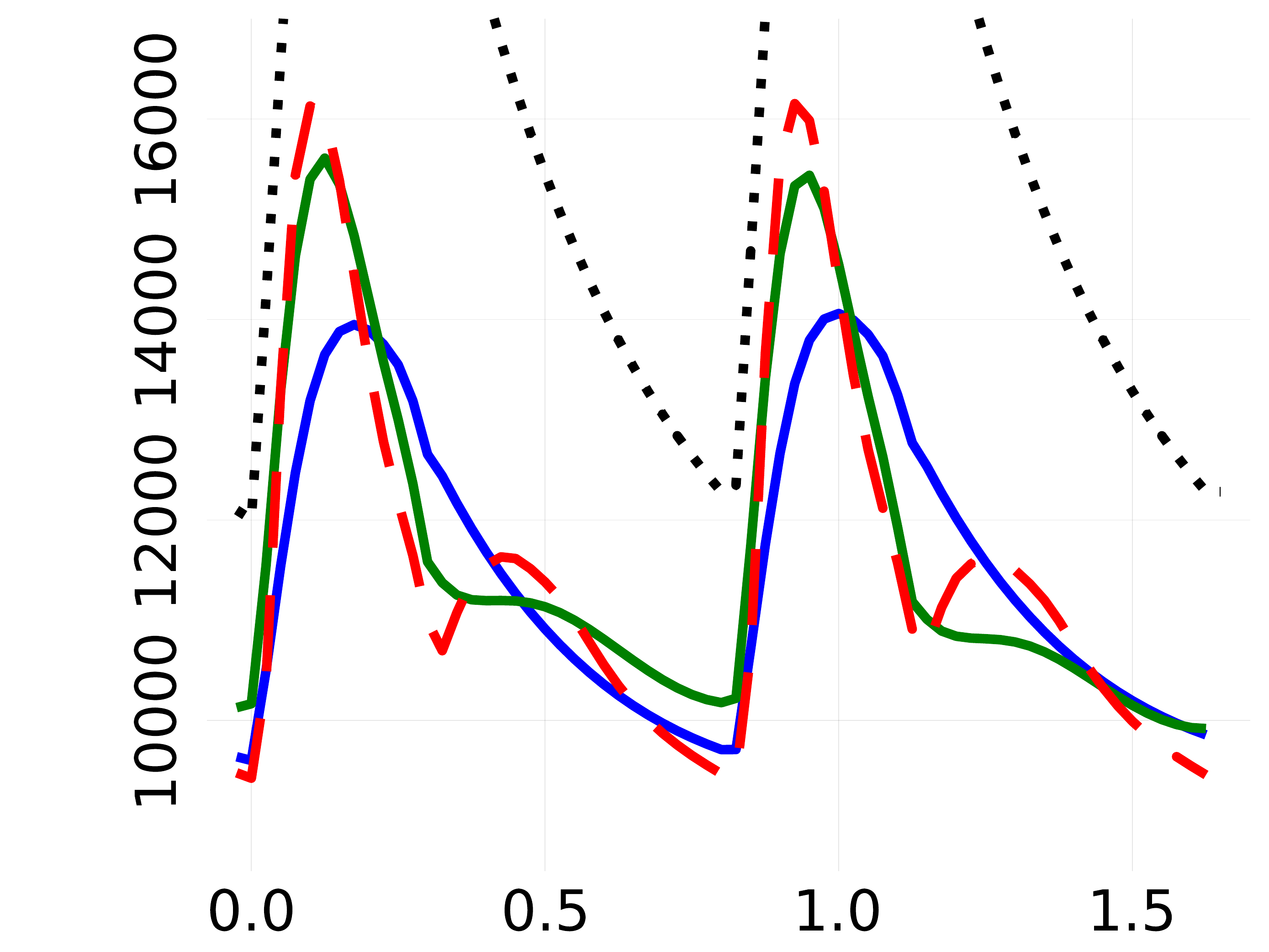}
		\caption{Digital Arteries (\#112)}
	\end{subfigure}
	\hfill
	\begin{subfigure}[b]{0.32\textwidth}
		\centering
		\includegraphics[width=\textwidth]{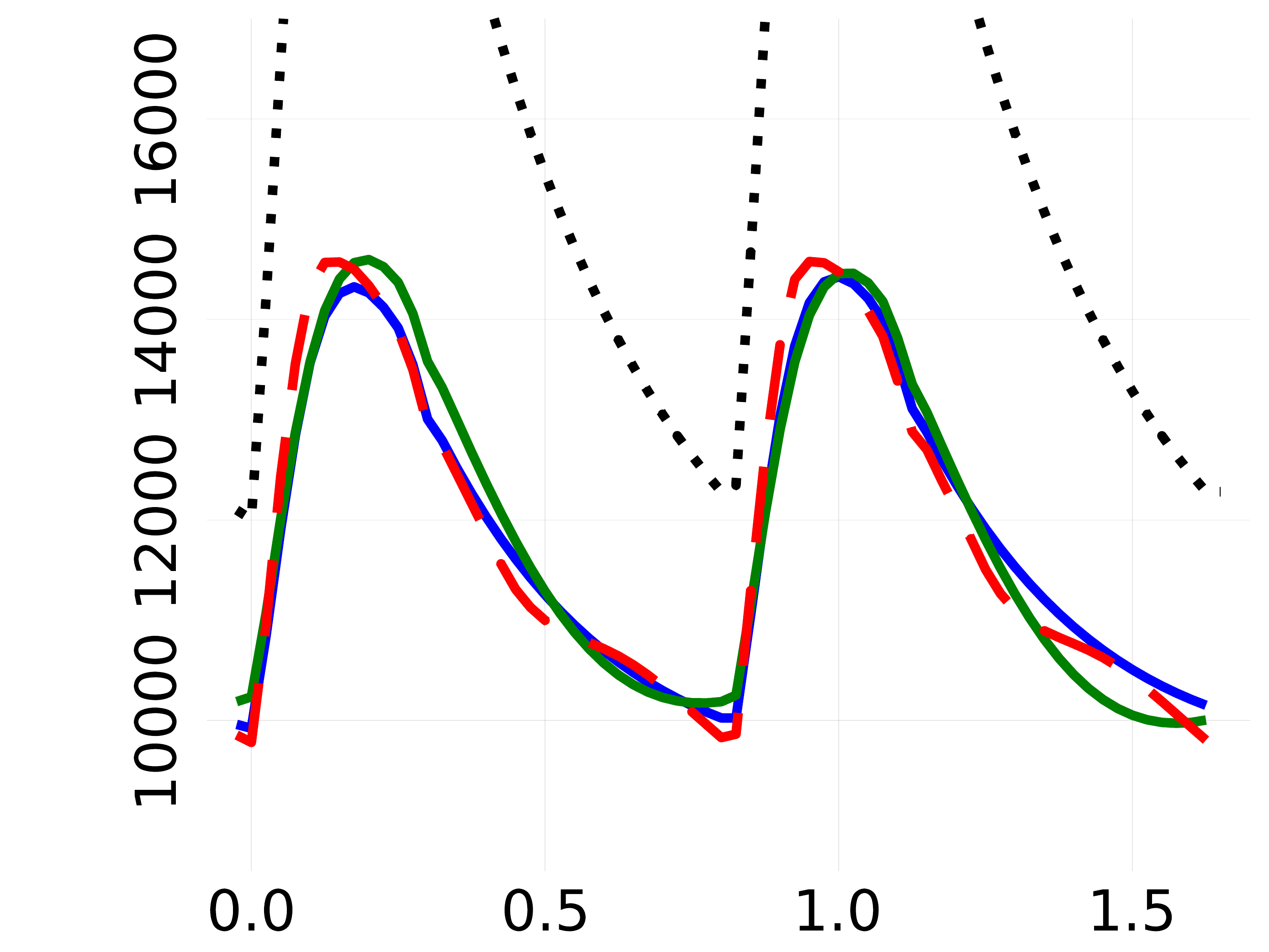}
		\caption{Com. Iliac Artery (\#44)}
	\end{subfigure}
	\hfill
	\begin{subfigure}[b]{0.32\textwidth}
		\centering
		\includegraphics[width=\textwidth]{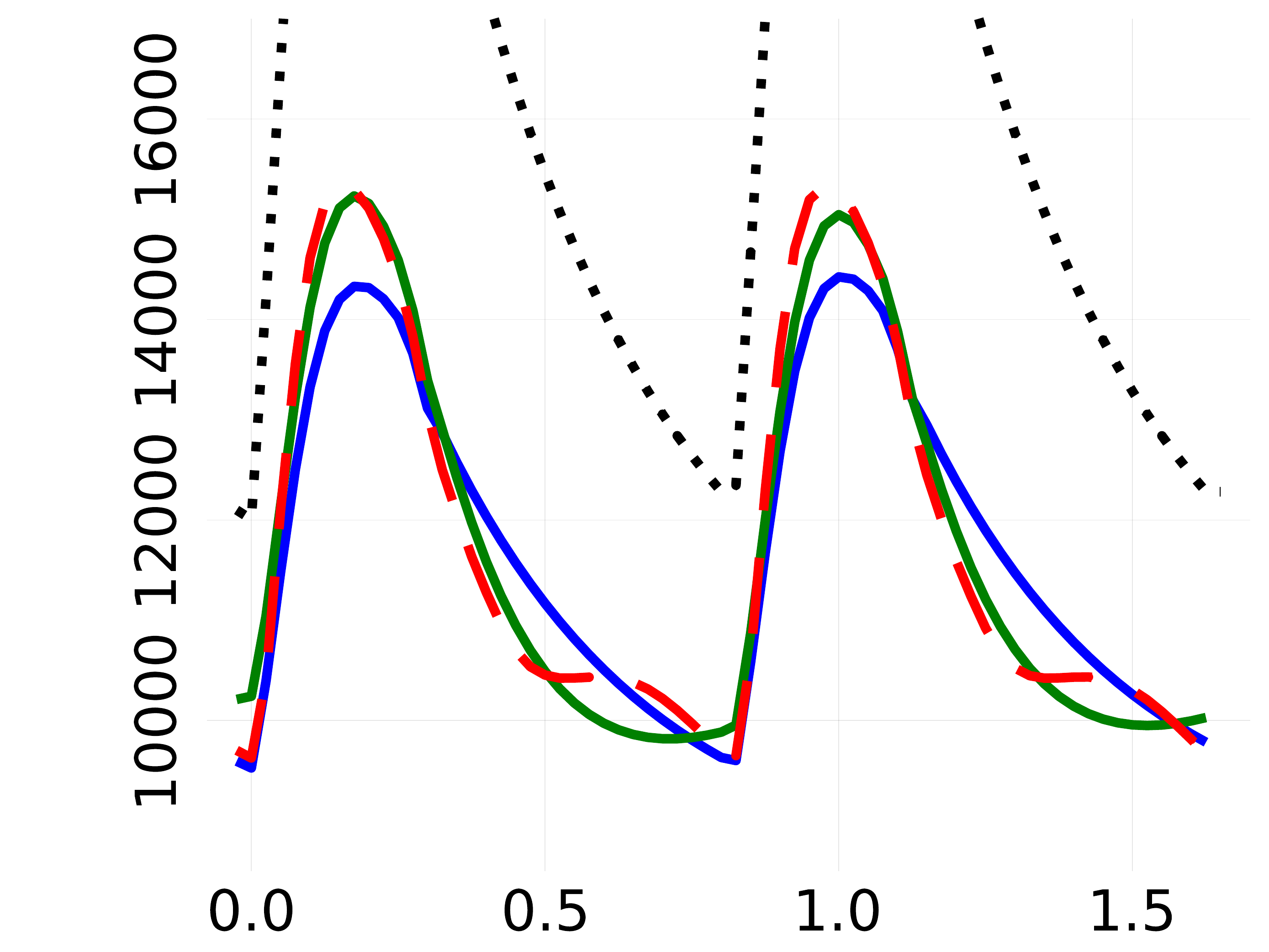}
		\caption{Femoral Artery (\#46)}
	\end{subfigure}
	
	\vspace{0.5cm}
	\begin{subfigure}[b]{0.32\textwidth}
		\centering
		\includegraphics[width=\textwidth]{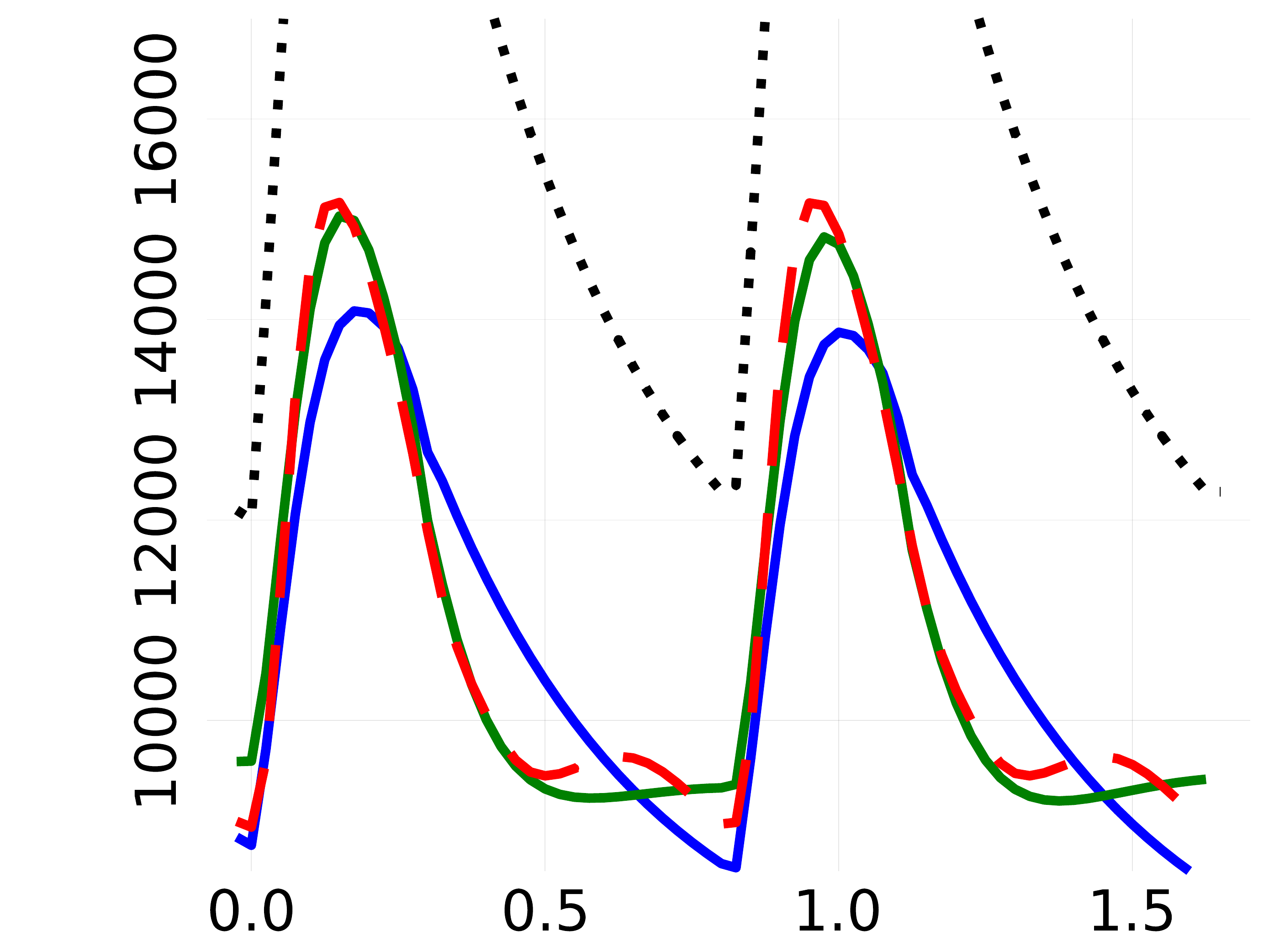}
		\caption{Ant. Tibial Artery (\#49)}
	\end{subfigure}

	\caption{Training results of the 10 considered arterial segments, based on data of subject \#1 of the \ac{PWDB} from \cite{Charlton:2019}. The horizontal axis labels the simulation time $t$ in seconds, the vertical axis the arterial pressure in Pascals. The plots show the pressure of the Modelica model (black/dotted, clipped), the target reference system pressure (red/dashed), the pressure learned by the NeuralFMU with C-placeholders (blue/solid) and with LC-placeholders (green/solid).}
	\label{fig:training}
\end{figure}

The training results, comparing the original \ac{FMU} pressure (from the Modelica model), the improved NeuralFMU pressures and the target pressure (reference system data) can be found in Fig. \ref{fig:training}. As loss function, a simple mean-squared-error with increasing time horizon between the reference system and Neural\ac{FMU} arterial blood pressures was deployed. The loss function rates the deviation only for a random sub-set of three artery segments, while changing the sub-set every training cycle (\emph{Batching}). The training itself was controlled via parameter freezing: Until reaching a significant small error (loss), parameters of the derivative \ac{ANN} (s. Fig. \ref{fig:structure}: bottom) were locked to force compensation of state offsets by the state \ac{ANN} (s. Fig. \ref{fig:structure}: top) and prevent early corruption of the system dynamics. After reaching a loss threshold, all net parameters were unlocked and trained in parallel.

For the C-placeholders, the static pressure wave from the FMU model is vertically scaled and shifted to fit the reference system data as good as possible. The LC-placeholders provide a better fit, because the LC-circuit additionally allows for phase-shifting the original pulse waves. The NeuralFMUs produce less data (only 10 dynamic locations instead of 116) and the output result does only approximate the reference model output, because of the relative small \ac{ANN} layout and a lack of states to interfere with the system. On the other hand, there is a significant gain in performance, which motivates the use of NeuralFMUs for use-cases, that focus on less measurement locations and/or more on fast computations than maximum precision, like embedded hardware or wearable computing. The reference model has a simulation time\footnote{AMD Ryzen\textsuperscript{TM} 9 3900X on Ubuntu 20.04.2 LTS} of $\approx 756\,s$ for 10 pulse waves, resulting in an average $\approx 75.6\,s$ per pulse waves. The Neural\acp{FMU} on the other hand, have a simulation time\footnote{Intel\textregistered{} Core\textsuperscript{TM} i7-8565U on Windows 10 Enterprise 20H2} of only $\approx 0.20\,s$ per 10 pulse waves, meaning an average of $\approx 0.02\,s$ per pulse wave\footnote{Even if the number of states and mathematical complexity of the \acp{FMU} differ between C- and LC-placeholders, the difference in simulation time of the resulting Neural\acp{FMU} was marginal.}. Even if both simulations were performed on different systems, a significant performance gain by a factor of $\approx 3750$ is clearly visible. This scope can be used to improve accuracy through deeper and/or wider network topologies or the injection of more and/or more complex state placeholders if necessary.

\begin{figure}[h!]
	\centering
	
	\begin{subfigure}[b]{0.32\textwidth}
		\centering
		\includegraphics[width=\textwidth]{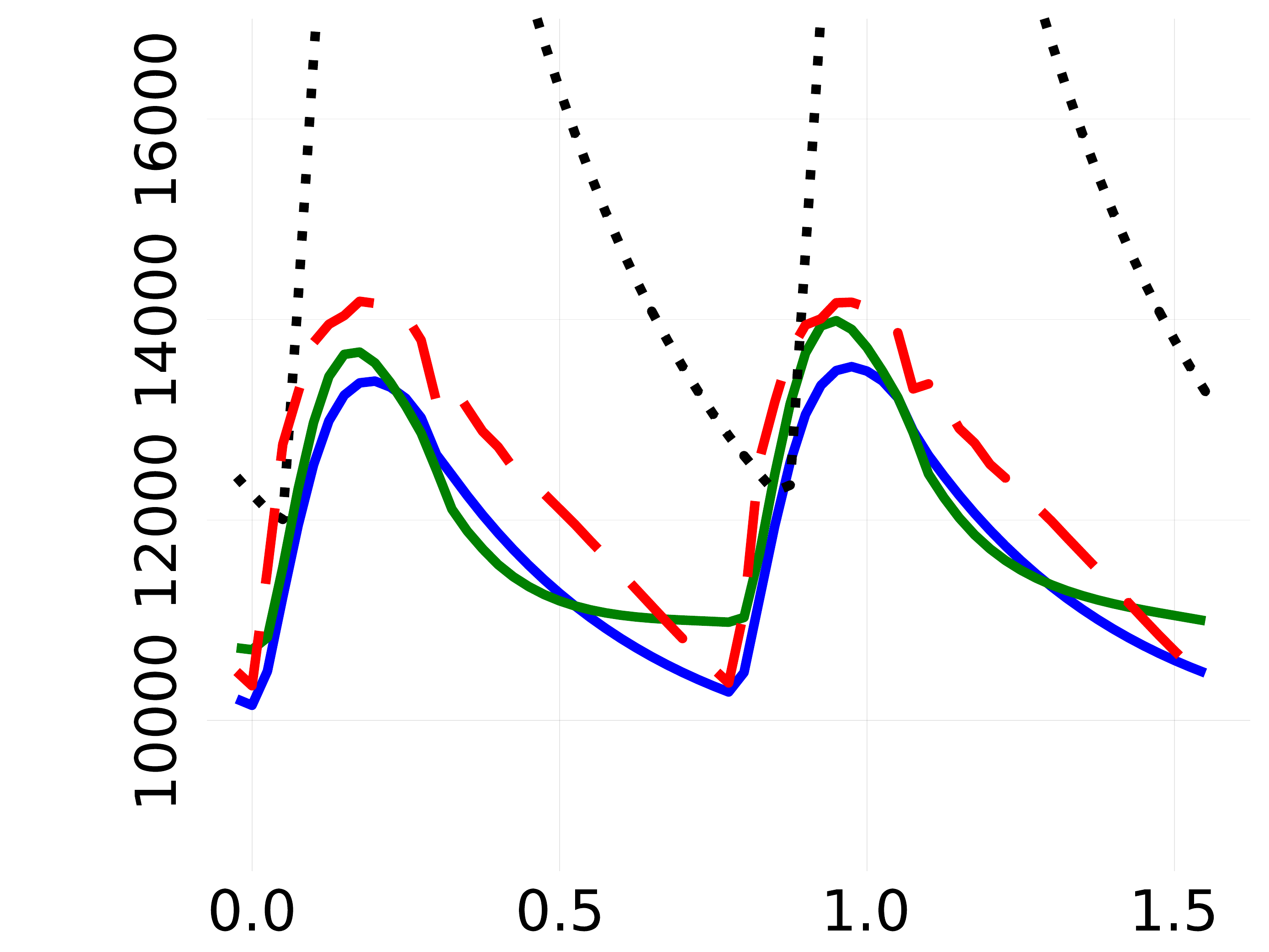}
		\caption{Thoracic Aorta (\#18)}
	\end{subfigure}
	\hfill
	\begin{subfigure}[b]{0.32\textwidth}
		\centering
		\includegraphics[width=\textwidth]{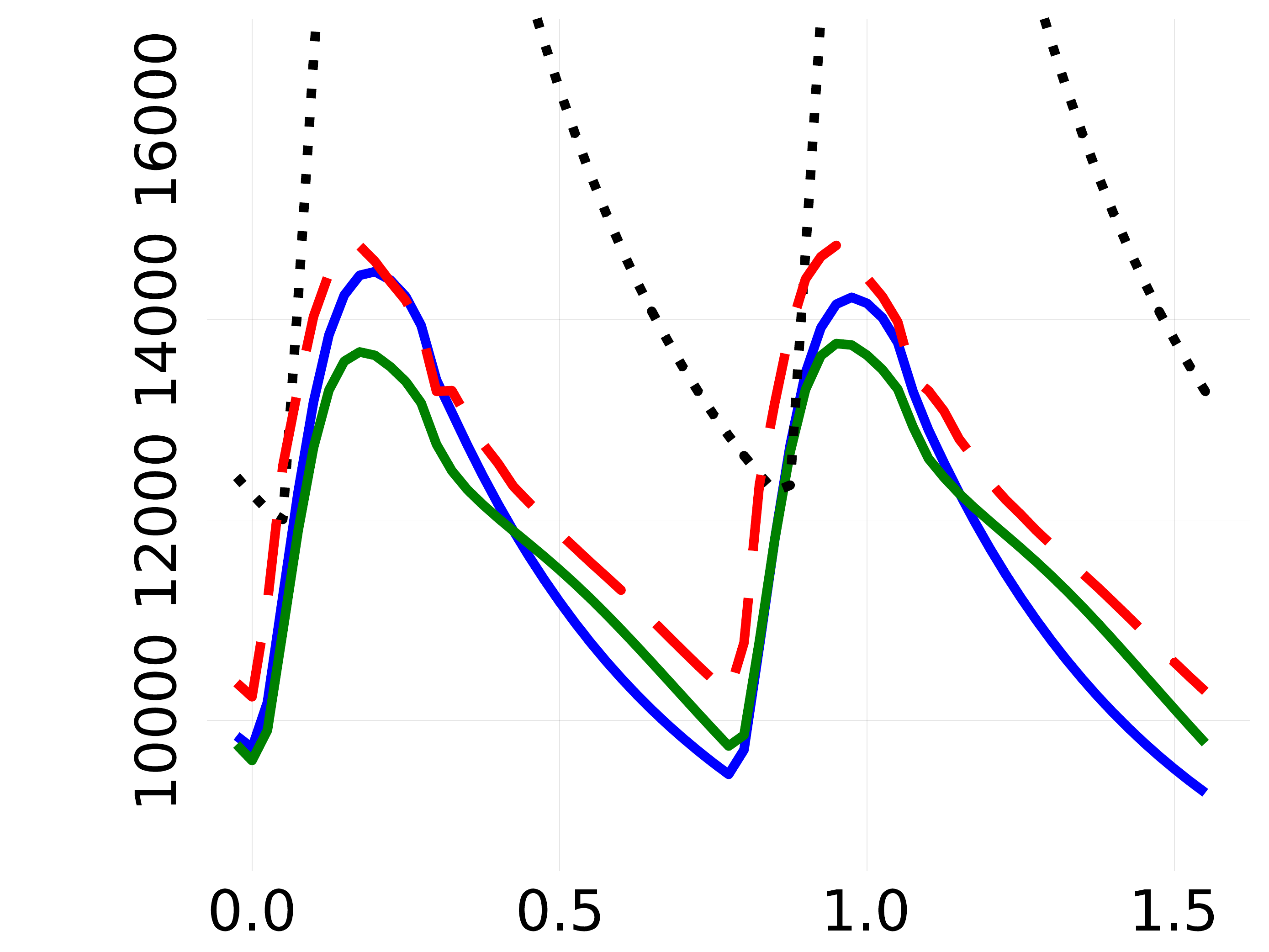}
		\caption{Abdominal Aorta (\#39)}
	\end{subfigure}
	\hfill
	\begin{subfigure}[b]{0.32\textwidth}
		\centering
		\includegraphics[width=\textwidth]{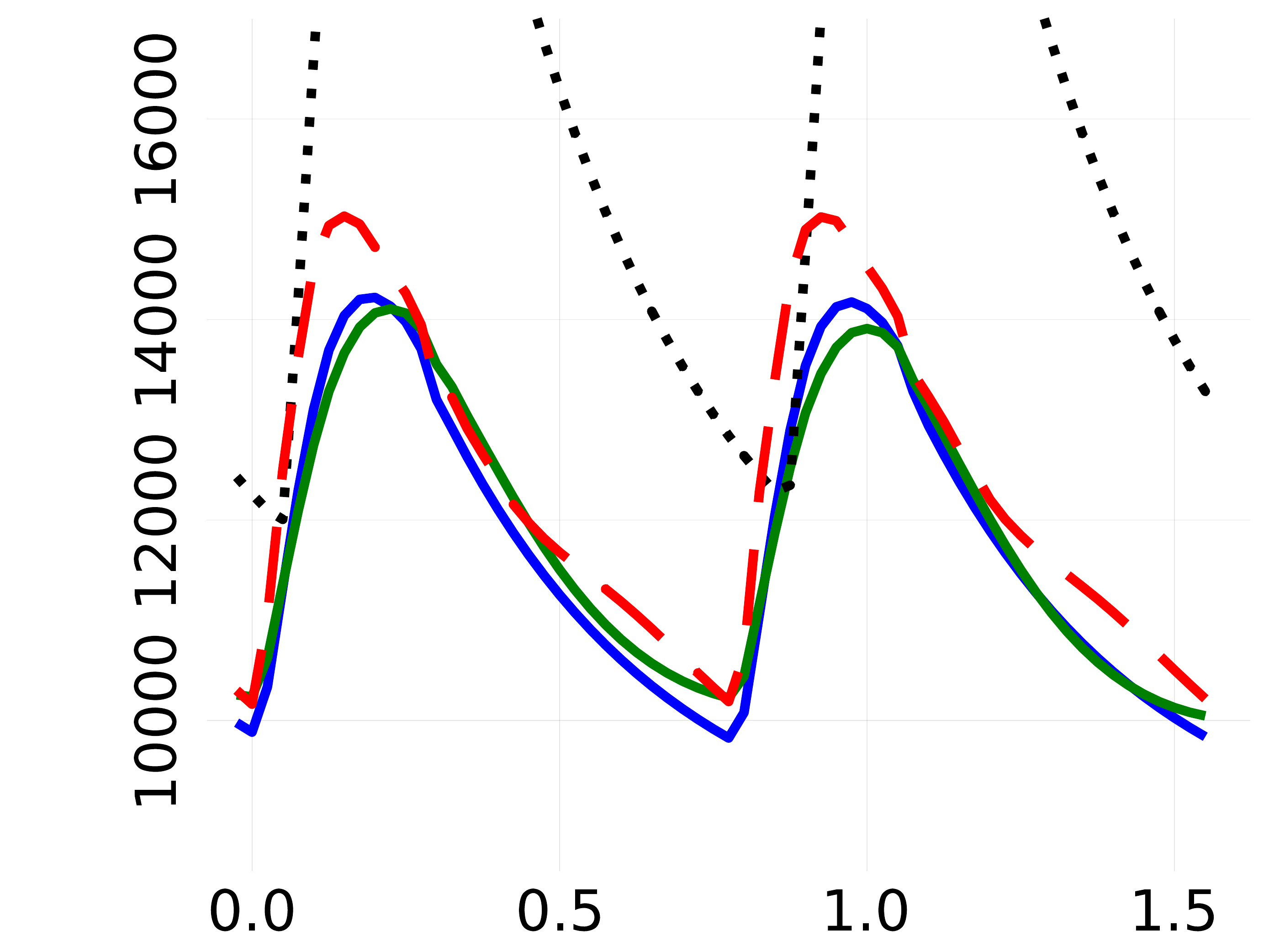}
		\caption{Iliac Bifurcation (\#41)}
		\label{fig:five over x}
	\end{subfigure}
	
	\vspace{0.5cm}
	\begin{subfigure}[b]{0.32\textwidth}
		\centering
		\includegraphics[width=\textwidth]{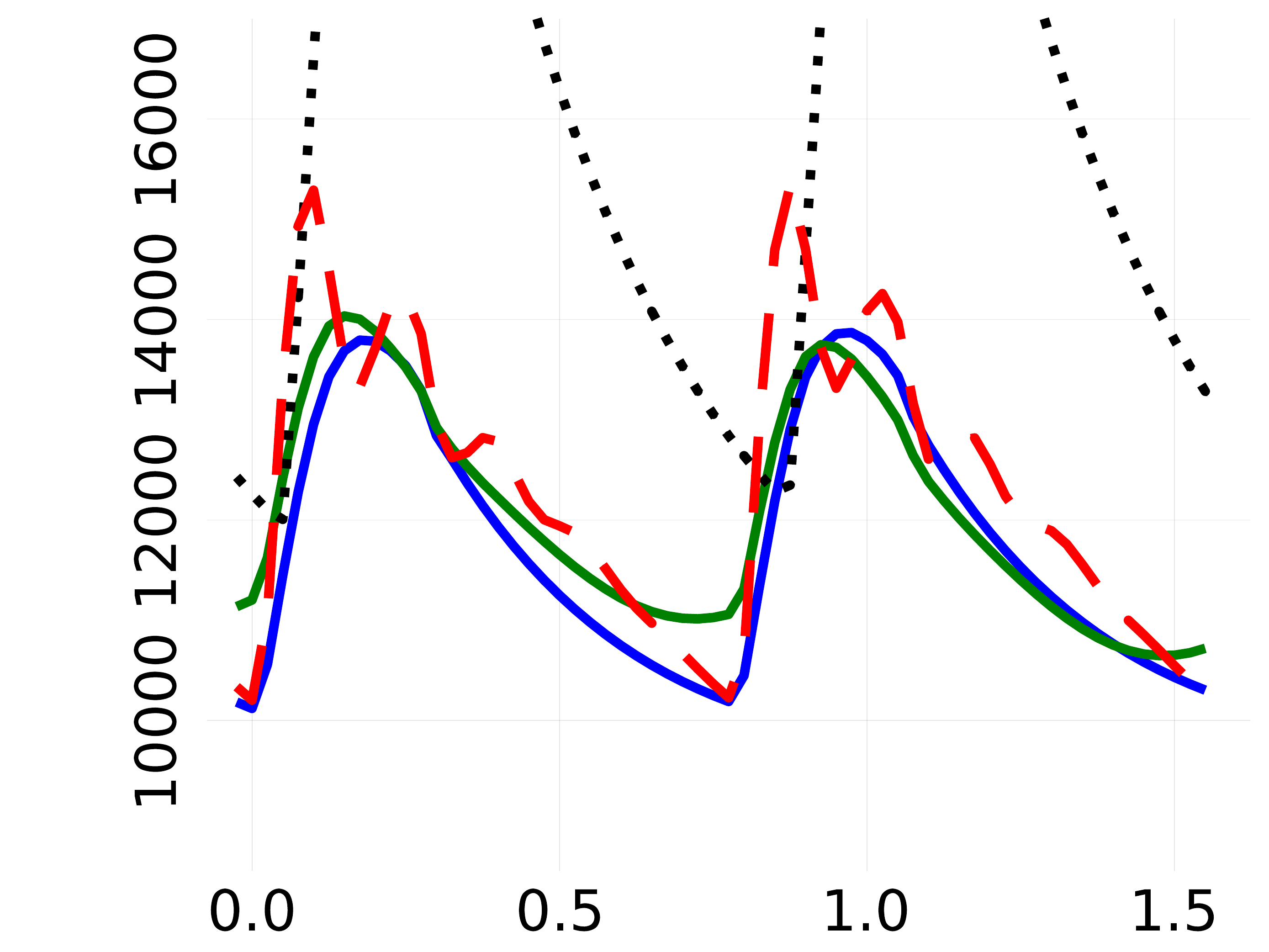}
		\caption{Sup. Temp. Artery (\#87)}
	\end{subfigure}
	\hfill
	\begin{subfigure}[b]{0.32\textwidth}
		\centering
		\includegraphics[width=\textwidth]{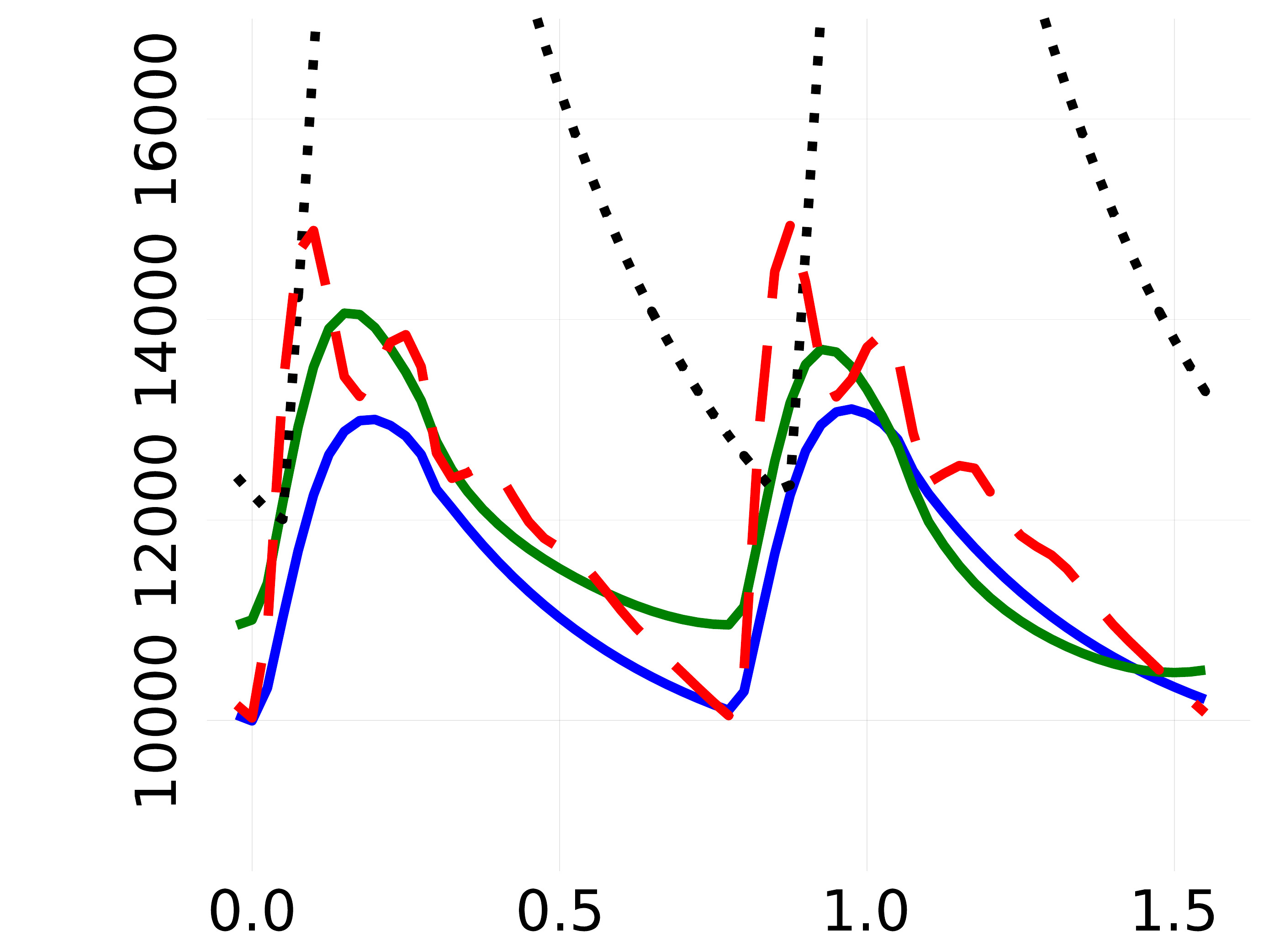}
		\caption{Mid. Cerebral Artery (\#72)}
	\end{subfigure}
	\hfill
	\begin{subfigure}[b]{0.32\textwidth}
		\centering
		\includegraphics[width=\textwidth]{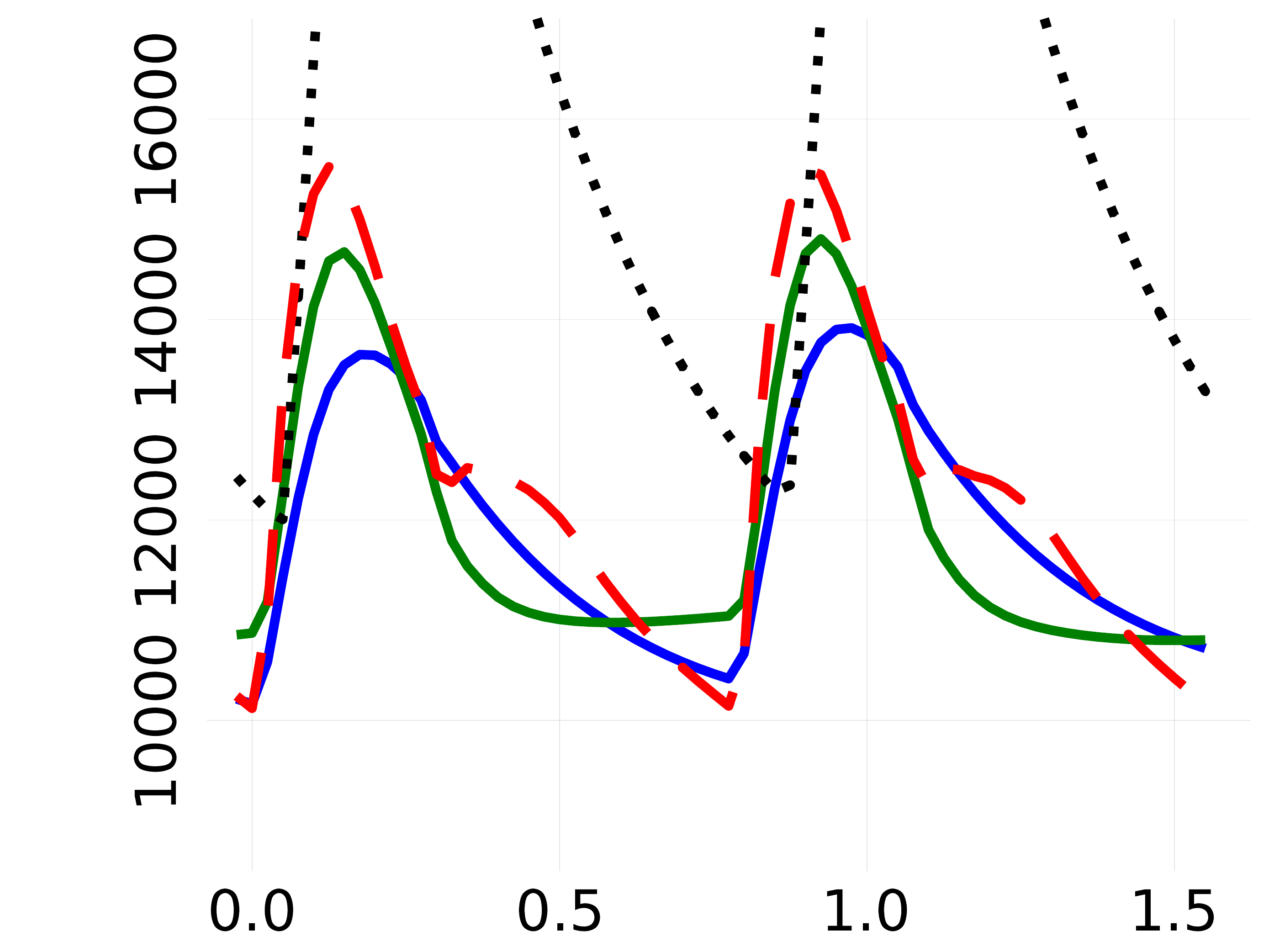}
		\caption{Brachial Artery (\#21)}
	\end{subfigure}
	
	\vspace{0.5cm}
	\begin{subfigure}[b]{0.32\textwidth}
		\centering
		\includegraphics[width=\textwidth]{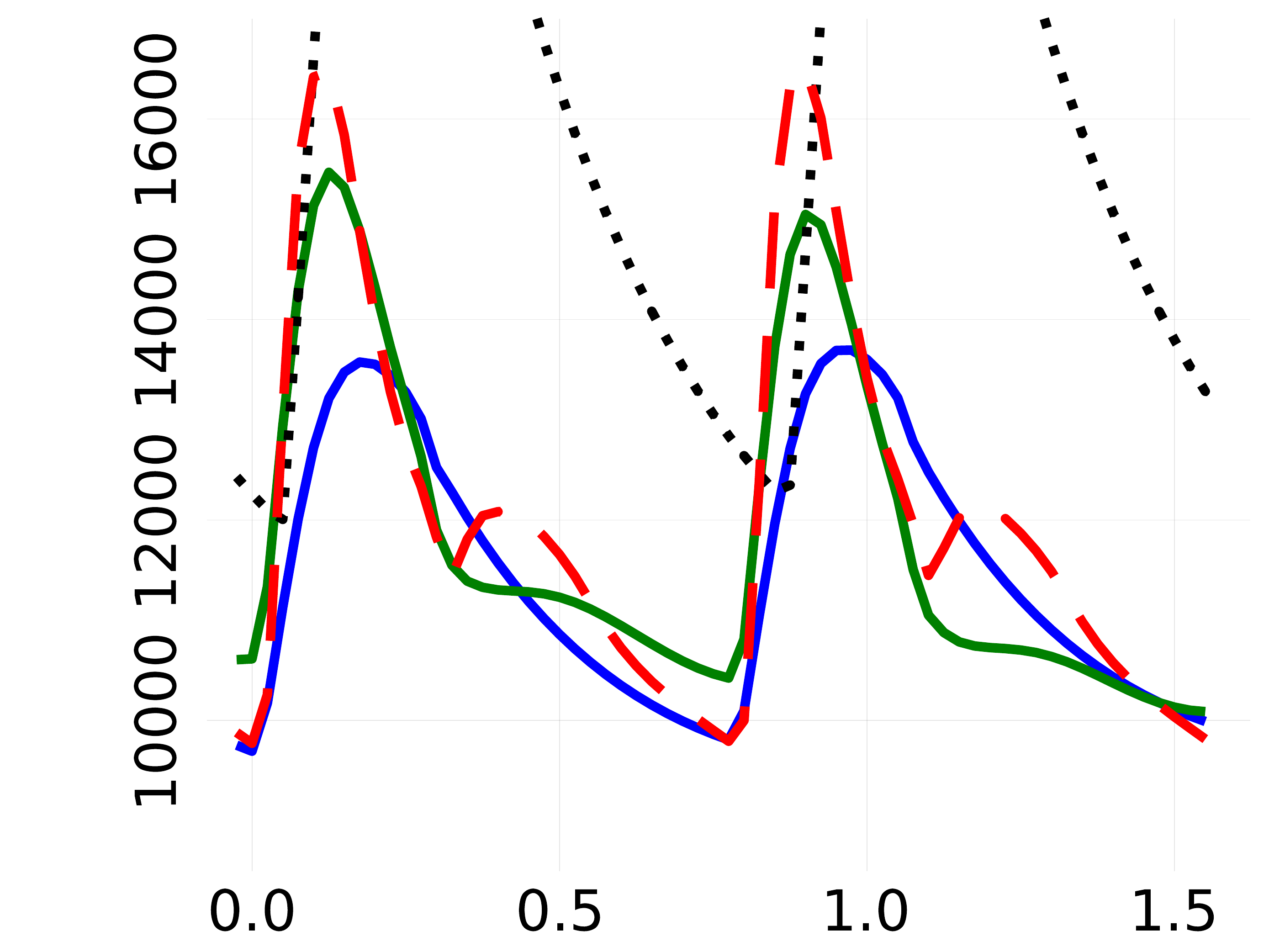}
		\caption{Digital Arteries (\#112)}
	\end{subfigure}
	\hfill
	\begin{subfigure}[b]{0.32\textwidth}
		\centering
		\includegraphics[width=\textwidth]{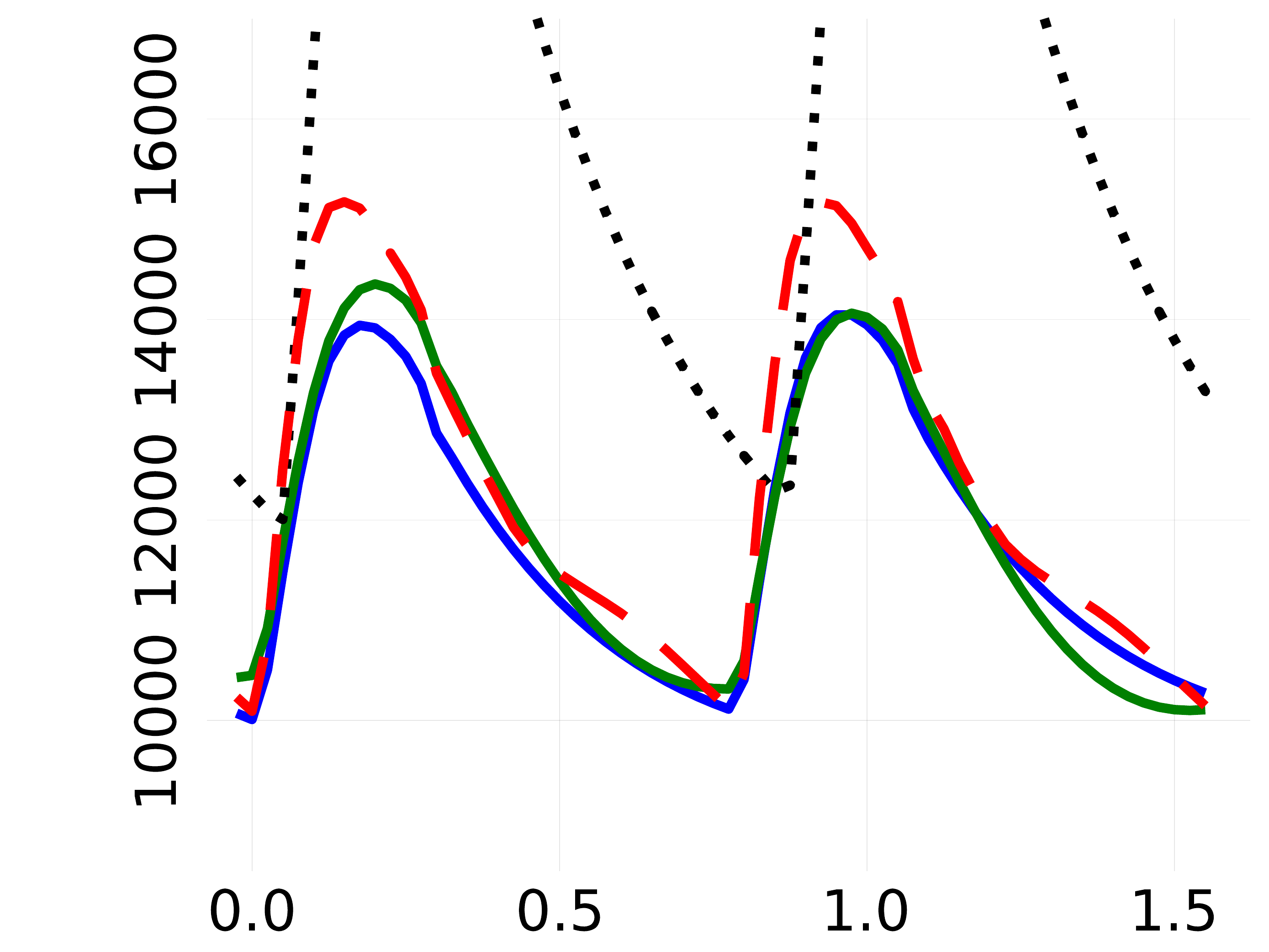}
		\caption{Com. Iliac Artery (\#44)}
	\end{subfigure}
	\hfill
	\begin{subfigure}[b]{0.32\textwidth}
		\centering
		\includegraphics[width=\textwidth]{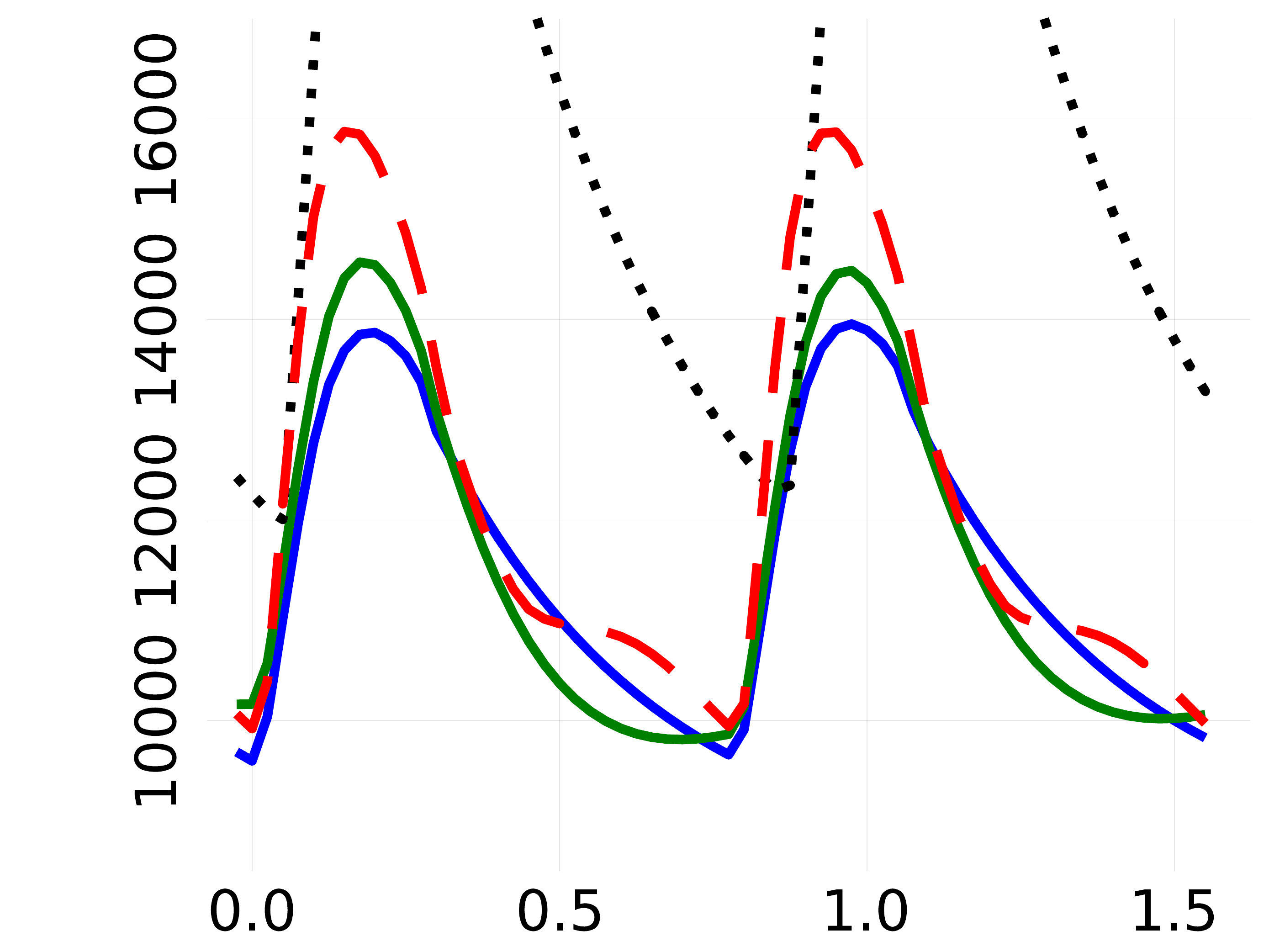}
		\caption{Femoral Artery (\#46)}
	\end{subfigure}
	
	\vspace{0.5cm}
	\begin{subfigure}[b]{0.32\textwidth}
		\centering
		\includegraphics[width=\textwidth]{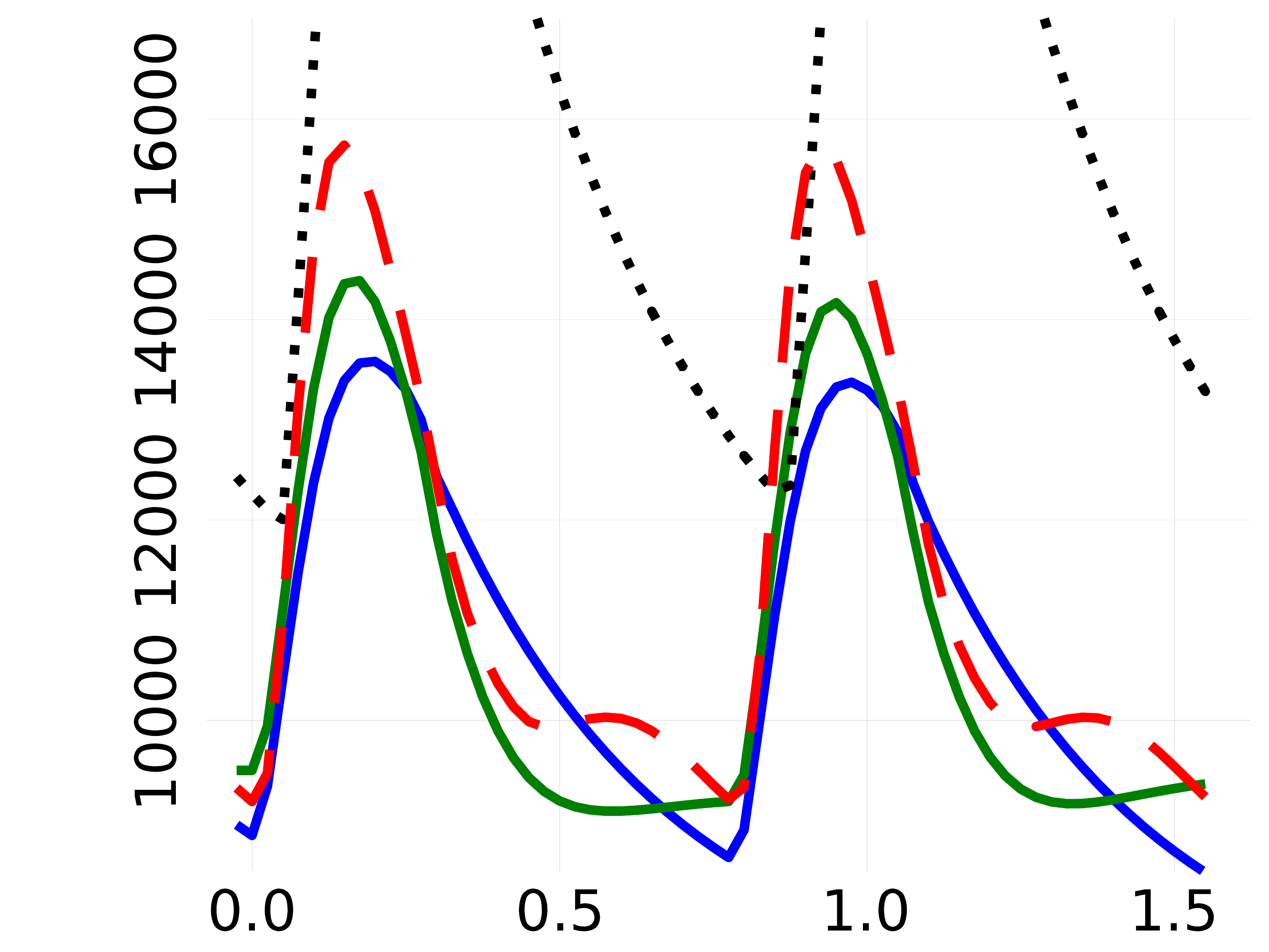}
		\caption{Ant. Tibial Artery (\#49)}
	\end{subfigure}
	
	\caption{Testing results on the 10 considered arterial segments. Tests where performed against data of (unknown) subject \#2 of the \ac{PWDB} from \cite{Charlton:2019}. The horizontal axis labels the simulation time $t$ in seconds, the vertical axis the arterial pressure in Pascals. The plots show the pressure of the Modelica model (black/dotted, clipped), the target reference system pressure (red/dashed), the pressure learned by the NeuralFMU with C-placeholders (blue/solid) and with LC-placeholders (green/solid).}
	\label{fig:testing}
\end{figure}

Testing the NeuralFMUs with unobserved data from patient \#2 leads to similar fitting pressure curves (s. Fig. \ref{fig:testing}), however the remaining average error is slightly bigger. Even if the derivative \ac{ANN} (bottom) did learn a simplified abstraction of the pressure dynamics, the simplification itself was understood in a generalized manner. Please note at this point, that the heart rate differs between the two patients ($\approx 73\,bpm$ for \#1 and $\approx 77\,bpm$ for \#2). The state \ac{ANN} (top) was only able to learn static offsets between the states of the reference system and the first principle model parameterized for patient \#1, but the offsets in data differ for patient \#2. It is visible in the plots, that a small vertical offset between training and testing results exists. These offsets could be reduced by training the state \ac{ANN} on data of patient \#2. 
A more detailed result validation is omitted at this point, but will be part of a pursuing contribution.

\section{Conclusion \& Future Work}\label{sec:conclusion}
We highlighted a workflow to improve a first principle model (at the example of a simple, object-orientated model) using a Neural\ac{FMU} without knowing the underlying physical equations necessary. The enhancements were learned solely based on a small set of data, in the presented example generated by a more accurate reference system. Starting with the \ac{FMI} model export from the modeling tool, the resulting \ac{FMU} was imported into the Julia programming language using the open-source library \libfmi{}\footnote{\urlfmi{}}. Inside Julia, a NeuralFMU was set up and trained, using the library extension \libfmiflux{}\footnote{\urlfmiflux{}}. Finally, the simulation data was compared to the target values from the \ac{PWDB}. The resulting Neural\ac{FMU} produces, dependent on the requirements of the underlying application, sufficiently accurate results. Model precision can be further improved by using more or other circuits for the state placeholders or wider and deeper \ac{ANN} structures. Further, the use of more than one patient during training will improve the Neural\ac{FMU} prediction quality on unknown patients. 

In terms of computational efficiency, a significant performance gain is visible for the considered example, making the presented approach interesting for performance critical applications. Neural\acp{FMU} have further advantages, e.g. even though they might include black-box models, they are fully differentiable. This allows for the use of other machine learning techniques or efficient model examination methods, like gradient based algorithms. 

Neural\acp{FMU} open up to many new interesting use-cases, but bring up many new challenges, too. For example, most physical systems are stable by default, meaning they converge against a physical equilibrium if not disturbed. By manipulating the system dynamics via an \ac{ANN}, this natural stability is not guaranteed anymore, which may result in a destabilized training process. Best practices for network initialization and stabilization methods during training must be considered and will be part of a pursuing contribution. 
\\
\\
The corresponding sources for the presented example will be published soon as part of the \libfmiflux{} library repository.

\section*{Acknowledgments}
This work has been partly supported by the ITEA 3 cluster programme for the project UPSIM - Unleash Potentials in Simulation.

\bibliography{research}

\end{document}